\documentclass[12pt]{alt2022} 

\title[PAC-Learning Uniform Ergodic Communicative Networks]{PAC-Learning Uniform Ergodic Communicative Networks}
\usepackage{times}
\usepackage[mathscr]{eucal}
\usepackage{amsmath,amssymb}
\usepackage{authblk}
\usepackage{mathtools}
\usepackage{algorithm,algorithmic}
\usepackage{booktabs} 
\usepackage{hyperref}
\usepackage{url}
\usepackage{natbib}
\usepackage{footnote}
\usepackage{graphicx}
\usepackage{bm}
\usepackage{paralist}
\usepackage{calrsfs}
\usepackage{subfiles}

\newtheorem{assumption}{Assumption}
\newtheorem*{lemma*}{Lemma}

\newtheorem{innercustomgeneric}{\customgenericname}
\providecommand{\customgenericname}{}
\newcommand{\newcustomtheorem}[2]{%
  \newenvironment{#1}[1]
  {%
   \renewcommand\customgenericname{#2}%
   \renewcommand\theinnercustomgeneric{##1}%
   \innercustomgeneric
  }
  {\endinnercustomgeneric}
}

\newcustomtheorem{customthm}{Theorem}
\newcustomtheorem{customlemma}{Lemma}

\newcommand{\bb}[1]{\mathbb{#1}}

\newcommand{\wh}[1]{\widehat{#1}}
\newcommand{\bfa}[1]{\boldsymbol{#1}}
\newcommand{\fk}[1]{\mathfrak{#1}}
\DeclareMathAlphabet{\pazocal}{OMS}{zplm}{m}{n}
\newcommand{\ca}[1]{\pazocal{#1}}
\newcommand{\mtt}[1]{\mathtt{#1}}
\newcommand{\pa}[1]{\mathcal{#1}}
\newcommand{\zt}{{Z_{\ca T}}}
\newcommand{\ezt}{{\ca Z_{\ca T}}}

\newcommand{\crad}[1]{{{\fk R}}_{\ca T}(#1)}
\newcommand{\ecrad}[1]{\wh{\fk R}_{\ca T}(#1)}
\newcommand{\orad}[1]{{\fk R}_{#1}(\ca H_{#1})}
\newcommand{\eorad}[1]{\wh{{\fk R}}_{#1}(\ca H_{#1})}
\newcommand{\lzt}[1]{_{\bfa L^{#1}(\zt)}}

\newcommand{\wrzt}{\widehat R_{Z_{\ca T}}(\bfa h)}

\newcommand{\vc}[1]{\mtt{vc}(\ca H_{#1})}
\newcommand{\erm}{\text{ERM}}

\DeclareMathOperator*{\argmin}{arg\,min}

\altauthor{%
 \Name{Yihan He} \Email{yihan.he@nyu.edu}\\
 \addr Courant Institute, NY, 10012
}

\begin{document}

\maketitle

\begin{abstract}%
  This work addressed the problem of learning a network with communication between vertices. The communication between vertices is presented in the form of perturbation on the measure. We studied the scenario where samples are drawn from a uniform ergodic Random Graph Process (RGPs for short), which provides a natural mathematical context for the problem of interest. For the binary classification problem, the result we obtained gives uniform learn-ability as the worst-case theoretical limits. We introduced the structural Rademacher complexity, which naturally fused into the VC theory to upperbound the first moment. With the martingale method and Marton's coupling, we establish the tail bound for uniform convergence and give consistency guarantee for empirical risk minimizer. The technique used in this work to obtain high probability bounds is of independent interest to other mixing processes with and without network structure.
\end{abstract}

\begin{keywords}%
  PAC-Learnability, Strong Mixing Process, Structured Prediction
\end{keywords}

\section{Introduction}

The problem of spatial statistical inference has great implications in the real world due to the large volume of geometric data today. In particular, the structured regression problem is of wide interest in signal processing and mathematical statistics. We consider the problem of structured prediction of labels in a network whose vertices communicate with each other frequently. This problem presents in different contexts, including epidemic on the network of person and social network inference based on history. In this problem, we are given a set of parties denoted as $V$ with $|V|$ parties. A subset of vertices will be activated and communicate with a fixed or varying neighborhood in each timestamp. The objective is to minimize the overall loss incurred by the miss-prediction on the label of activated vertices across a period of time. Related problems have been widely studied both by computer scientists ~\citep{birant2007st} and statisticians~\citep{alkon1988spatial,handcock1994approach} in history, which is often referred to as spatial temporal inference.

We build the problem with the assumption that the data is sampled according to a random graph process. The samples generated by an RGP formed a sequence of networks indexed by their timestamps where the sample generated by a party is represented by a unique vertex, and the instant communication between vertices is broadcasted along the edges. Parties might be active (when they present in the network) or passive (when they are simply isolated vertices). A key idea here is that the communication to a vertex by its neighborhood will affect the distribution of its samples in the upcoming timestamp. The regularity is then characterized by the dependence of measure, or \textit{ mixing conditions }. For some random process $Z:(Z_k:k\in\bb Z)$, we say a real-valued process is strongly mixing if $$\bfa P(Z_i=z_1,Z_{i+n}=z_2)\overset{n\to\infty}{\longrightarrow}\bfa P(Z_i=z_1)\bfa P(Z_{i+n}=z_2)$$ uniformly for $z_1,z_2\in\bb R$ and $i\in\bb Z$. In other words, the two random variables generated by the same process that stays far away from each other are asymptotically independent of each other. 

To formalize the learning problem, at time slot $t$ we activate a set of vertices $\ca V_t\subset V$ with a graph $\ca G_t$ with $\ca E_t$ being its edge set. The vertex set $\ca V_t$ contains $N_t$ vertices indexed by their index in $V$. Coupled with the timestamp, vertex indexed jointly by $(t,i)$ has its own feature and label written as $X_{t,i}\in\bb X\subset\bb R^d$ and $Y_{t,i}\in\bb Y\subset\bb R$. We are given a sequence of hypothesis sets $\{\ca H_1,\ldots,\ca H_T\}$ adaptively or deterministically and the loss function $L:\bb Y\times\bb Y\rightarrow \bb R^+$. Hence, the incurred average loss over the active set $\ca V_t$ is $R_t=\frac{1}{N_t}\sum_{i\in\ca V_t}L(h_t(X_{t,i}),Y_{t,i})$ over the whole network. We simplify the pair of feature/label as $Z_{t,i}=(X_{t,i},Y_{t,i})$ for some $(t,i)$. The final loss over the whole network from time $1$ to $T$ is written as $\frac{1}{T}\sum_{t=1}^TR_t$.

It is important to note the difference between this problem with adversarial online learning. In the latter problem, no consistency is expected from the learner. The best we can achieve is the sub-linear regret, while in this work, we established uniform learn-ability. In other words, we focused on the worst-case limits. When i.i.d. assumption is no longer present because of the network communication, the learn-ability will depend on the mixing condition of networks. We also gave guarantees under the non-communicative assumptions as a comparison, where the 
party in the network is assumed to be independent of each other.

The technical difficulties come in two folds: \textbf{(1)} The expected worst-case loss with the dependence measure. To address this, we designed the structural Rademacher complexity, which yields a finer bound than the sequential Rademacher complexity proposed in~\citet{rakhlin2010online}. The structural Rademacher complexity can naturally fit into VC theory by chaining. It is estimated as $\ca O\bigg (\sqrt{\frac{1}{T\inf_{t} N_t}\mtt{vc}(\pa H)}\bigg )$ where $\mtt{vc}(\pa H)$ is the VC dimension of a product hypothesis set. The technique applied here is dimension reduction on the concentrated mixing measure. \textbf{(2)} A non-vacuous upper bound on the tail for the dependent measure. The idea to address this problem comes from the martingale method~\citep{kontorovich2007measure} and Marton Coupling~\citep{marton2004measure}. To conclude, the problem we studied relies heavily on two regularity conditions: The rate of mixing ( or mixing time ) and the sparsity of the communication graph. For a strongly mixing process with an exponential rate of convergence and relatively sparse neighborhood, our bound is non-vacuous. Given this bound, we are able to obtain consistency guarantees of the empirical risk minimizer.

\section{Related Works}
The problem of learning RGP is mainly interlaced with three strings of work. The first one is adversarial online learning with a multi-agent communicative network, or cooperative learning pioneered by~\citet{awerbuch2008competitive}. In this problem, agents are represented as vertices in a fixed structured network and communicate with their neighbors in the learning process to jointly solve an online learning problem. Each agent represents a learner with its model updated independently from others. The more recent development of this line includes ~\citet{cesa2016delay,rubiodecentralized,cesa2020cooperative,della2021efficient}, where they studied various cooperative online learning problems and gave efficient algorithms with asynchronous updates. This work is different from theirs as we focus on the structured prediction of networks, and each vertex in the network is considered as a sample in the learning problem instead of a learner. On the other hand, we not only focus on regret minimization but on establishing uniform learn-able classes and procedures like ERM. This can be seen through that our bound is high probability instead of the first moment.

The second line of work focuses on the measure concentration without independent assumptions. Previously, many attempts have been made to the asymptotics of mixing processes. The mixing condition is a broad literature that involves many different variants of requirements, which has been summarized in ~\citet{}. In particular, \citet{yu1994rates} proved asymptotic rates of uniform convergence under stationary $\beta$-mixing processes. Other works also addressed the asymptotic rates of $\alpha$-mixing, $\phi$-mixing and $\eta$-mixing ~\citep{vidyasagar2013learning,kontorovich2008concentration,kontorovich2007measure}. ~\citet{adams2010uniform} proved the convergence of VC class under ergodic processes without the use of symmetrization. 
The method in this work is inspired partially by the martingale methods, but the problem is fundamentally different from this line of work in the learning context.

The third string aims at developing the metric entropies and combinatorics to give guarantees of regret in online learning problems. A steady line of work by ~\citet{ littlestone1988learning,ben2009agnostic,shalev2014understanding} studies the online learnability via combinatorial dimensions like Littlestone's and give guarantees for the realizable case through the depth of mistake tree.  This line of work is further extended by~\citet{rakhlin2010online,rakhlin2011online,rakhlin2015sequential}, which introduced the sequential complexities and sequential covering numbers. Those combinatorics takes the maximum over the value of the process. Hence, their result gives an upper bound on the adversarial learnability in the form of the first moment of regret. However, as the learnability, if non-stochastic online learning guarantees no consistency, their bound will be pessimistic in our problem. Our introduction of structural Rademacher complexity generalizes the sequential process into networks. On the other hand, our guarantees will be strictly better than the sequential complexities as the expectation is always a lower bound on the supremum in the mixing case.  

\section{Preliminaries}

\subsection{Random Graph Process (RGP) }
In the previous literature ~\citep{van2017stochastic}, the random graph process is a sequence of graphs with random structure on a fixed vertex set. This is different from the one studied in this work, where we focus on the random feature in the networks. For a random graph process $\ca Z_{\ca T}$ defined on the domain $\bb Z=\bb X\times\bb Y$ as the product of feature and label. $\ca Z_{\ca T}$ hence induces a product state space $\Omega_1^T=\Omega_1\times\cdots\times \Omega_T$. We can draw a sample consisting of $T$ sets of feature/label pair from $\Omega_1^T$ according to $\ca Z_{\ca T}$ and denote the graphs as $\ca G_1^T$. Each vertex can be active (when it stays within the network) or inactive (when it is isolated from other vertices) in $\ca G_t$. We also note that a bijection from $\pi_t:[N_t]\to\ca V_t$ can be defined, which maps the re-indexed index to the active indices in $V$ at timestamp $t$. The active set of vertices is denoted by $\ca V_t$ and the set of all undirected edges connecting two non-identical vertices are denoted as $\ca E_t$. Vertex $i$ at timestamp $t$ is indexed by $(t,i)$ and is given a random feature label pair $(X_{t,i}, Y_{t,i})$ that is simplified as $Z_{t,j}$. This form $T$ batches of samples denoted by $Z_{\ca T}=\pa Z_1^T=\{\pa Z_1,...,\pa Z_T\}$ where we denote $\pa Z_t=\{Z_{t,j}:j\in\ca V_t\}\subset\bb Z$ and $\ca T $ as the universal vertex set 
of $\ca Z_{\ca T}$. We simplify the notations by denoting  $\pa X_{t}=\{X_{t,j}:j\in \ca V_t\}$, $\pa Y_{t}=\{Y_{t,j}:j\in \ca V_t\}$, $\pa X_i^j=\{\pa X_i,...,\pa X_j\}$,  $\pa Y_i^j=\{\pa Y_i,...,\pa Y_j\}$, $\pa Z_i^j=\{\pa Z_i,...,\pa Z_j\}$ and $\Omega_i^j=\prod_{k=i}^j\Omega_k$. We assumed that $\pa X_1^0=\pa Z_1^0=\pa Z_0=\pa X_0=\emptyset$.

Then we introduced several assumptions on the dependence of measure. The first assumption is on the limits of communication. One can see that the perturbation from a single vertex will broadcast along the edges until the whole network receives it. 

\begin{assumption}
Given a graph $\ca G_t(\ca V_t,\ca E_t)$ with vertex set $\ca V_t$ and unordered edge set $\ca E_t$ generated at timestamp $t$. Denote the neighborhood of vertex $i$ in $\ca G_t$ by $\ca N_{i,t}=\{j: (i,j)\in\ca E\}$ and the feature/label pair in the neighborhood as $Z_{\ca N_{i,t}}=\{Z_{k,t}:k\in\ca N_{i,t}\}$. Then the following holds
\begin{align*}
    \bfa P(Z_{t+1,i}|\pa Z_1^{t})=\bfa P(Z_{t+1,i}|\pa Z_1^{t-1}, Z_{\ca N_{i,t}})
\end{align*}
\end{assumption}


The second assumption is that two vertices within the same timestamp correlated with each other only because they received the message from the same vertex. This assumption is   referred to as the Local Markov assumption in graphical models. We put here for technical reason.
\begin{assumption}[Conditional Independence]
Given $i,j\in V$, we assume that \begin{align*}
    \bfa P(Z_{t,i},Z_{t,j}|\pa Z_1^{t-1})=\bfa P(Z_{t,i}|\pa Z_1^{t-1})\bfa P(Z_{t,j}|\pa Z_1^{t-1})
\end{align*} for all $t \in [T]$.
\end{assumption}
This assumption implies that the feature/label pair of vertices in the same timestamp is independent upon observing their common ancestors.

\subsection{Ergodic and Independent RGP}
We are interested in the effect of communicative vertices in the network learning problem. To approach it mathematically, we assumed an exponential rate of mixing on the dependence measure. ( also referred to as uniform ergodicity in the context of Markov chains ) This condition implies strong mixing in the context of strictly stationary Markov chains. Intuitively, if we pick a single path of message broadcast in the network, the \textit{message} decays along its way. We also discuss its degenerated case where communication is eliminated from the process. 
\begin{definition}[Uniform Ergodic RGP]
An RGP with vertex set $V$ has stationary distribution $\pi$ for all vertices, state space $\Omega_i$ for $\pa Z_i$ is uniform ergodic if its convergence to the stationary measure is geometric e.g. for all $z\in\prod_{i=1}^{t-1}\Omega_i$, we have:
\begin{align*}
    \sup_{z\in\Omega_1^i}TV(P(Z_{t,j}|\pa Z_1^i=z),\pi)\leq\frac{k_0}{N_t}\rho^{t-i}
\end{align*}
where $\rho\in (0,1)$, $k\in\bb R^+$, and  $TV$ is the total variation difference defined by $
    TV(\mu,\nu):=\sup_{A\in\ca F}\bigg| \mu(A)-\nu(A)\bigg|
$ for measure $\mu,\nu$ on $\sigma$ algebra $\ca F$.



\end{definition}
\begin{remark}
The assumption of uniform ergodicity is consistent with the one in univariate stochastic processes. The $\frac{1}{N}$ term implies that uniform ergodic RGP is also a multivariate stochastic process defined by $\pa Z_1^T$. This $\frac{1}{N}$ coefficient also helps us to establish the learnability, which we will find out in later sections.
\end{remark}
We used the martingale method~\citep{kontorovich2007measure} and derived the learning guarantee of ergodic RGP with Marton Coupling~\citep{marton2004measure}. The notations by ~\citet{paulin2015concentration} is presented here for completeness.
\begin{definition}[Marton Coupling]
Let $\pa Z_1^T:=(\pa Z_1,...,\pa Z_T)$ be a vector of random variables taking values in $\Omega_1^T=\Omega_1\times...\times\Omega_T$. A Marton's coupling for $\pa Z_1^T$ is a set of couplings:
\begin{equation*}
    (\pa Z^{(z_1,...,z_i,z_i^\prime)},\pa Z^{\prime (z_1,...,z_i,z^\prime_i)})\in\Omega_1^T\times\Omega_1^T
\end{equation*}
for every $i\in[T]$, every $z_1\in\Omega_1,...,z_i\in\Omega_i,z_i^\prime\in\Omega_i$, satisfying the following conditions:
\begin{compactenum}
    \item$
        \pa Z_1^{(z_1,...,z_i,z^\prime_i)}=z_1, ...,\pa Z_i^{(z_1,...,z_i,z_i^\prime)}=z_i,\;\;\;\;
        \pa Z_1^{\prime (z_1,...,z_i,z_i^\prime)}=z_1,...,\pa Z_i^{\prime (z_1,...,z_i,z_i^\prime)}=z_i^\prime,
   $ 
    \item\begin{align*}
        (\pa Z_{i+1}^{(z_1,...,z_i,z^\prime_i)},....,\pa Z_T^{(z_1,...,z_i,z^\prime_i)})&\sim \bfa P(\pa Z_{i+1},...,\pa Z_T|\pa Z_1=z_1,...,\pa Z_i=z_i)\\
        (\pa Z_{i+1}^{\prime (z_1,...,z_i,z^\prime_i)},....,\pa Z_T^{\prime (z_1,...,z_i,z^\prime_i)})&\sim \bfa P(\pa Z_{i+1},...,\pa Z_T|\pa Z_1=z_1,...,\pa Z_i=z_i^{\prime })
    \end{align*}
    \item If $z_i=z_i^\prime$, then $\pa Z^{(z_1,...,z_i,z_i^\prime)}=\pa Z^{\prime (z_1,...,z_i,z_i^\prime)}$
\end{compactenum}
\end{definition} 
And the following notion of mixing matrix is defined, which is a slight modification from the one in~\citet{marton2004measure}.
\begin{definition}[Mixing Matrix]
For an RGP $\ezt$, we define its upper triangular mixing matrix $\bfa \Gamma$. When $k>t$, we have
\begin{align*}
    \Gamma_{t,k}=
     \sum_{j\in \ca V_t} \sup_{z_1^t, z_{t,j},z_{t,j}^\prime}TV(\bfa P(Z_{k,j}|\pa Z_1^t=z_1^t,Z_{t,j}=z_{t,j}),\bfa P(Z_{k,j}|\pa Z_1^t=z_1^t,Z_{t,j}=z_{t,j}^\prime)) 
\end{align*}
when $k=t$, we have
\begin{align*}
    \Gamma_{t,k}=1
\end{align*}
\end{definition}

\subsection{Learning Theory}
We introduce several notations that is used through out this work. The learning process incurred the following empirical and generalization error.
\begin{equation*}
     \widehat R_{Z_{\ca T}}(\bfa h)=\sum_{t=1}^T\sum_{j\in \ca V_t}\frac{1}{TN_t} L(h_t(X_{t,j}),Y_{t,j})\;\;\text{, and}\;\;R_{\ca Z_{\ca T}}(\bfa h)=\bb E_{ Z_{\ca T}\sim \ca Z_{\ca T}}\widehat{R}_{Z_{\ca T}}(\bfa h)
\end{equation*}
We will use $\bb E_{\ezt}$ to denote $\bb E_{\zt\sim\ezt}$, that is, the expectation over the process. Similar to the online learning problem, we can measure the algorithm's performance through regret, which is represented as the loss on the hindsight (empirical and expected). The empirical and expected regret is written as
\begin{equation*}
    \widehat {\ca R}_{Z_{\ca T}}(\bfa h)=\wrzt-\inf_{\bfa h\in\pa H}\wh R_{Z_{\ca T}}(\bfa h)\;\;\text{, and}\;\;\ca R_{\ca Z_{\ca T}}(\bfa h)=R_{\ezt}(\bfa h)-\inf_{\bfa h\in\pa H}R_{\ca Z_{\ca T}}(\bfa h)
\end{equation*}

\section{Combinatorial Dimensions}
To approach the learnability of uniform ergodic RGPs, we need to give an upper bound on the first moment of empirical processes. Typically, symmetrization and chaining is the standard method to approach its upper bound. However, those method does not directly apply to the problem of mixing processes. As a result, we proposed novel combinatorics termed \textit{Structural Rademacher complexity} that is suitable for the problem of interest.
In the sequel, we present the main results with occasionally the proof sketch but delay detailed technicalities to the appendix for interested readers.
\subsection{Rademacher Complexity}

Other types of combinatorics are also introduced for the purpose of upper-bounding the regret of online learning in the adversarial setting. Typically, 
a string of previous work by ~\citet{littlestone1988learning,ben2010theory,ben2009agnostic,rakhlin2010online, rakhlin2015sequential} introduced sequential Rademacher process and complexities to give upper bound on the regret of online learning. Their argument takes supremum over all possible trees valued functions. In the problem of upper bounding mixing processes, their method is over-pessimistic. By switching the supremum argument to the conditional expectation, we can thus obtain a finer bound for the mixing processes.

We first review the standard form of Rademacher complexity and then introduce \textit{structural Rademacher complexities}, followed by an upper bound for it via metric entropies.

\begin{definition}[Structural Rademacher Complexities]
For a sequence of hypothesis sets $\{\ca H_1,\ldots\ca H_T\}$ and $\zt$ sampled according to $\ezt$ with index space $\ca T$. Let $ \sigma_{t,j}\in\{\pm 1\}$ with $(t,j)\in\ca T$ be a set of i.i.d. Rademacher random variables, 
For the set consisting of all $T$-dimensional hypothesis sequence $\pa H=\{(h_1,...,h_T):h_t\in\ca H_t\text{ for all } t\in[T]\}$, the \textbf{structural Rademacher complexity} and its empirical variant are defined by:
\begin{align*}
    &\crad{\pa H}=(\orad{1},...,\orad{T})^\top,\text{ with }\;\orad{t}=\bb E_{\ezt,\bfa\sigma}\bigg[\sup_{h\in\ca H_t}\sum_{j\in \ca V_t}\frac{\sigma_{t,j}}{TN_t}h(X_{t,j})\bigg|\pa X_1^{t-1}\bigg]\\
    &\ecrad{\pa H}=(\eorad{1},...,\eorad{T})^\top,\;\;\text{ with }\;\eorad{t}=\bb E_{\bfa\sigma}\bigg [\sup_{h\in\ca H_t}\sum_{j\in \ca V_t}\frac{\sigma_{t,j}}{TN_t}h(X_{t,j})\bigg]
\end{align*}
\end{definition}
Intuitively, the structural Rademacher complexity constructs a coupling for each timestamp and can be seen as a by-product of the RGP.

Following ~\citet{ledoux2013probability}, we give the contraction lemma for the structural Rademacher complexity. 
\begin{lemma}[Contraction]
\label{2}\label{contract}
Let $\Phi_1,...,\Phi_T$ be $l$-Lipschitz functions of $\bb R\rightarrow\bb R$ and $\bfa \sigma$  be the i.i.d. Rademacher random field with index space $\ca T$. 
    Then, for the structural Rademacher complexity, the following holds:
    \begin{equation*}
        \ecrad{\Phi_t\circ\ca H_t}=\sup_{h\in\ca H_t}\frac{1}{TN_t}\sum_{j\in \ca V_t}\sigma_{t,j}(\Phi_t\circ h)(X_{t,j})\leq l\ecrad{\pa H}_t\;\;\text{ for all } t\in[T]
    \end{equation*}
\end{lemma}
 
The structural Rademacher complexity connects with its empirical counterpart through the following:
\begin{lemma}\label{lm:7}
Let $\pa H=\{(h_1,...,h_T):h_t\in\ca H_t,\forall t\in[T]\}$ be a hypothesis set for $T$-dimensional hypotheses such that any hypothesis $h\in\ca H_i$ takes value in $[-M,+M]$ for all $i\in[T]$. Let $Z_{\ca T}$ be an RGP sampled according to $\ca Z_{\ca T}$, then with probability at least $1-\delta$ for $\delta\in(0,1)$ the following holds:
\begin{align*}
    \crad{\pa H}_t\leq\ecrad{\pa H}_t +\frac{M}{T}\sqrt{\frac{1}{2N_t}\log\frac{1}{\delta}},\;\;\Vert\crad{\pa H}\Vert_1\leq\Vert\ecrad{\pa H}\Vert_1 +\frac{M}{T}\sqrt{\frac{\log\frac{T}{\delta}}{2}}\sum_{t=1}^T\frac{1}{\sqrt{N_t}}
\end{align*}
In particular, if $\zt$ is independent, we have with probability at least $1-\delta$:
\begin{equation*}
    \bb E\Vert\crad{\pa H}\Vert_1\leq\Vert\ecrad{\pa H}\Vert_1+\frac{M}{T}\sqrt{\sum_{t=1}^T\frac{1}{2N_t}\log\frac{1}{\delta}}
\end{equation*}
\end{lemma}

\begin{remark}
The standard Rademacher complexity does not have similar results when $\ca Z_{\ca T}$ is a general RGP. Our design of structural Rademacher complexity addressed this problem.
\end{remark}
\subsection{Metric Entropy}
\label{rdm}
Structural Rademacher complexities can also connect with the standard metric entropies via chaining. Here we first review the standard definition of covering and packing in the function family. 
\begin{definition}[Covering Number]
A set $N$ is called an $\epsilon$-net for function space $\ca H$ with pseudometric $d$ (e.g. $(\ca H,d)$) if for every $h\in \ca H$, there exists $\pi(h)\in N$ such that $d(h,\pi(h))\leq\epsilon$. The smallest cardinality of an $\epsilon$-net for $N$ is called the covering number, denoted as
$
    N(\ca H,d,\epsilon):=\inf\{|N|:\text{ N is an $\epsilon$-net for $(\ca H,d)$}\}
$
\end{definition}
\begin{definition}[Packing Number]
 A set $N\subseteq \ca H$ ia called an $\epsilon$-packing of $(\ca H, d)$ if $d(h ,h^\prime)\geq \epsilon$ for every $h,h^\prime\in N, h\neq h^\prime$. The largest cardinality of an $\epsilon$-packing of $(\ca H, d)$ is called the packing number, denoted as
$
     D(\ca H,d,\epsilon):=\sup\{|N|:\text{ N is an }\epsilon\text{-packing of }(\ca H, d)\}
$
\end{definition}
\begin{lemma}
For every $\epsilon>0$:
$
    D(\ca H,d,2\epsilon)\leq N(\ca H,d,\epsilon)\leq D(\ca H,d,\epsilon)
$
\end{lemma}
Extended from ~\citep{dudley1984course}, the following bound hold for the Rademacher complexities of RGP:
\begin{theorem}[Chaining]\label{chaining} For a $\zt$ sampled according to $\ezt$, the following holds for empirical structural Rademacher complexities
\label{5}
 \begin{align*}
      \Vert \ecrad{\pa H}\Vert_1\leq C\sqrt{\frac{1}{T\inf_tN_t}}\int_0^\infty\sqrt{\log N(\pa H, d_{\bfa L^2(\zt)}, \epsilon)}d\epsilon
 \end{align*}
 with
 $
     d_{\bfa L^2(\zt)}(\bfa f,\bfa g)=\bigg[\sum_{t=1}^T\sum_{j\in [N_t]}\frac{1}{TN_t}\big (f_t(X_{t,j})-g_t(X_{t,j})\big )^2\bigg]^{\frac{1}{2}}\\
$ and $C$ being a universal constant.
\end{theorem}

The following lemma binds covering numbers of different metrics for $\pa H$.
\begin{lemma}
\label{6}
\label{cov}
\label{lm:13}
$ N(\pa H,d\lzt{2},\epsilon)\leq N(\pa H,d\lzt{\infty},\epsilon)\leq N(\pa H,d_{\infty},\epsilon)
$ with
\begin{equation*}
     d_{\bfa L^\infty(\zt)}(\bfa f,\bfa g)=\sup_{t,j}\big |f_t(X_{t,j})-g_t(X_{t,j})\big |\;\text{, and }\;d_{ L^\infty}(\bfa f,\bfa g)=\sup_{t\in[T],x\in\bb X}\big |f_t(x)-g_t(x)\big |
\end{equation*}
\end{lemma}

\subsection{VC Combinatorics}
In this section, we give new VC-combinatorics for the Boolean valued function family equipped with the state space of a random graph process. Readers familiar with the context of adversarial online learning might remind combinatorics like Littlestone dimensions~\citep{littlestone1988learning} that guarantee the error of deterministic strategy. This line of work is different from this one since the bound given by those adversarial dimensions doesn't guarantee consistency.

Our main results include an extension of Sauer's lemma and an almost dimension-free upper bound on the covering number in $d_{\bfa L^2(\zt)}$ metric. Intuitively, when the number of active vertices is not steady, say it grows consistently with the number of timestamps, the hypothesis class used for the first timestamp might not be powerful enough to shatter all vertices at time $T$. Thus we will need a larger class of functions.

We first extend the definition of shattering, VC dimensions, and growth functions for RGPs. Then we discuss the connections of these combinatorics with the metric entropies introduced in previous sections, followed by the statement of the final dimension-free upper bound for them. This bound gives upper bound on the first moment error rate in $\ca O\bigg (\sqrt{\frac{1}{T\inf_tN_t}\tt vc(\pa H)}\bigg )$. 

Let $\ca C_t$ be the concept class (A set of subspaces) on $\bb X$ and $\pa H=\{(\bfa 1_{C_1},...,\bfa 1_{C_T}):C_t\in\ca C_t\text{ for all } t\in[T]\}$ be the hypothesis set of $T$-dimensional binary hypotheses induced by $\pa C=(\ca C_1,...,\ca C_T)$. We extend the notion of shattering~\citep{vapnik2013nature} to the state space of RGPs.

\begin{definition}[Shattering an RGP]
A RGP $Z_{\ca T}\in\Omega_1^T$ is said to be shattered by a sequence of concept classes $\ca C_1,...\ca C_T$ (or shattered by $\pa C$ ) if  
\begin{equation*}
    |\ca C_t\cap  \pa X_t| =2^{N_t} \;\;\text{ for all } t\in [T]
\end{equation*}
then for any subset $\tau\subseteq \ca V_t$, denote the corresponding subset of $\pa Z_t$ by $\pa Z_{\tau}=\{Z_{i}: i\in\tau\}$ and $\pa X_{\tau}=\{X_{i}:i\in\tau\}$, there exists $C\in\ca C_t$ such that $C\cap \pa X_t=\pa X_\tau$. $\Pi_{\pa H}(Z_{\ca T})=\prod_{t=1}^T|\ca C_t\cap \pa X_t|$ is also called \textbf{Generalized Growth Function} of $\pa H$ on $\zt$.
\end{definition}
We give a VC dimension extension directly from shattering
\begin{definition}[VC-dimension]
\label{def:8}
The VC dimension of concept class $\pa C$ on $\bb Z$ is defined by 
$\mtt{vc}(\pa C):=\sup\{|\ca T|=\sum_{t=1}^TN_t: \text{There exists }\;Z_{\ca T}\in\Omega_1^T\text{ that is shattered by }(\ca C_1,..,\ca C_T) \}$
\end{definition}
We then immediately found the connection between the standard VC dimensions of concept classes and its RGP variant.

\begin{lemma}\label{lm:16}
When the RGP is independent, $\mtt{vc}(\pa C)=\sum_{i=1}^T\mtt{vc}(\ca C_i)$, otherwise $\mtt{vc}(\pa C)\leq\sum_{i=1}^T\mtt{vc}(\ca C_i)$
\end{lemma}

Then we use the generalized growth function to upper bound the covering number in Lemma \ref{cov}
\begin{definition}[Covering Number of Concept Classes]
Using $N(\pa C,d,\epsilon)$ to denote the $\epsilon$-covering number of a $T$-dimensional concept class $\pa C=\{(C_1,...,C_T):C_i\in\ca C_i\}$ with the $L^p(\zt)$ norm defined by $d_{L^p(\zt)}(\bfa C,\bfa C^\prime)=\big [\sum_{t=1}^T\sum_{i\in[N_t]}\frac{1}{TN_t}(\bfa 1_{ C_t}(X_{t,i})-\bfa 1_{ C_t^\prime}(X_{t,i}))^p\big]^{1/p}$ where $Z_{\ca T} $ is an RGP and $\bfa C,\bfa C^\prime\in\pa C$.  By definition, we have $N(\pa C,d,\epsilon)=N(\pa H,d,\epsilon)$.
\end{definition}
The following lemma connects covering number with the value of growth function for the same function family
\begin{lemma}
\label{lm:18}
\label{8}
\label{dual2}When $\epsilon\leq 1$, we have
$
    N(\pa H,d_{L^\infty(Z_{\ca T})},\epsilon)=\Pi_{\pa H}(Z_{\ca T}), \text{ and } N(\pa H,d_{\infty},\epsilon)=|\pa C|
$
\end{lemma}

We then give the counterpart of Sauer's lemma for RGP, 
\begin{lemma}[Sauer's Lemma]\label{lm:19}
\label{11}
For $Z_{\ca T}\in\Omega_1^T$ and hypothesis set $\pa H$ with index space $\ca T$, the growth function is upper-bounded by:
\begin{equation*}
    \Pi_{\pa H}(Z_{\ca T})\leq\prod_{t=1}^T\sum_{k=0}^{\min\{\mtt{vc}(\ca C_t), N_t\}}\binom{N_t}{k}\leq\prod_{t=1}^T\bigg(\frac{e N_t}{\min\{\mtt{vc}(\ca C_t), N_t\}}\bigg)^{\min\{\mtt{vc}(\ca C_t), N_t\}}
\end{equation*}
\end{lemma}

\begin{remark}
For the general RGP, the upper bound is sharp. When $N_t=\mtt{vc}(\ca C_t)$ and assume that no set of cardinality greater than $N_t$ can be shattered by $\ca C_t$ and all subsets of cardinality $\mtt{vc}(\ca C_t)$ can be shattered, the first inequality will become equality.
\end{remark}
We note that the upper bound is still vacuous as the sample size $N$ of each timestamp becomes large. This bound is the best we achieved without extra assumptions on the communication pattern. Then we give a finer upper bound for independent and uniform ergodic processes. Our technique to obtain upper bounds is probabilistic extraction~\citep{dudley1978central} on the mixing measure. We start with the independent/non-communicative regime.
\begin{lemma}
\label{lm121}
Assuming that $\zt\in\Omega_1^T$ is sampled according to a non-communicative $\ezt$. Let $f_1,..., f_m$ be the functions on $\zt$ such that
\begin{equation*}
    \Vert \bfa f_i\Vert_\infty\leq 1,\;\;\;\;\Vert \bfa f_i-\bfa f_j\Vert_{\bb E\bfa L^2(\zt)}\geq\epsilon\;\;\;\;\text{for all }1\leq i<j\leq m
\end{equation*}
Or in other words, $\bfa f_1,\ldots,\bfa f_m $ formed a $\epsilon$-net on $\pa H$.
Then there exist a $z_{\ca T}$ sampled according to $\ezt$ with $\frac{T^2}{\sum_{t=1}^T\frac{1}{N_t}}>\frac{16}{9}\epsilon^{-4}\log m$ such that:
\begin{equation*}
    d_{\bfa L^2(z_{\ca T})}(\bfa f,\bfa g)=\Vert\bfa f_{i} -\bfa f_{j}\Vert_{
\bfa L^2(z_{\ca T})}\geq \frac{\epsilon}{2}\;\;\;\; \text{for all }1\leq i<j\leq m,
\end{equation*}
with
$
    \Vert\bfa f_i-\bfa f_j\Vert_{\bb E\bfa L^2(Z_{\ca T})}=d_{\bb E\bfa L^2(\zt)}(\bfa f,\bfa g)=\bigg(\bb E_{\zt}\bigg[\sum_{t=1}^T\sum_{j\in [N_t]}\frac{1}{TN_t}\big (f_t(X_{t,j})-g_t(X_{t,j})\big )^2\bigg]\bigg)^{\frac{1}{2}}\\
$ be the expected metric.
\label{lm:21}
\end{lemma}

Then it comes to the upper bound for covering number. The idea is proving measure-independent upper bound.
\begin{theorem}
\label{13}
\label{thm:22}
When the $\zt\in\Omega_1^T$ is sampled according to a non-communicative $\ezt$, for all $\epsilon\in (0,1]$ we have:
\begin{align*}
     \log N(\pa H,d_{\bb E\bfa L^2(\ezt)},\epsilon)&\leq\prod_{t=1}^T\bigg(\frac{32e}{9T\epsilon^4}\bigg)^{2\vc t}\\
     \Vert\ecrad{\pa H}\Vert_1&=\ca O\bigg(\sqrt{\frac{1}{T|V|}\mtt{vc} (\pa H)}\bigg)
\end{align*}      

\end{theorem}
 
Then we move to the uniform ergodic process. It is important to note the technicalities involved when moving from the concentration of independent to mixing measure. This follows the similar martingale method and coupling construction.
\begin{lemma}
\label{lm122}
\label{lm:23}
Assume that $\zt\in\Omega_1^T$ is sampled according to ergodic $\ezt$. Let $\bfa f_1,...,\bfa f_m$ be the functions with the same conditions of lemma \ref{lm121}.
Then there exists a $z_{\ca T}$ sampled according to $\ezt$ with  $\frac{1}{\sum_{t=1}^T\frac{1}{N_t^2}}\geq\frac{16\Vert\bfa\Gamma(\ca T)\Vert_2^2}{9}T^{-4}\epsilon^{-4}\log\frac{m(m-1)}{2}$ such that:
\begin{equation*}
    d_{\bfa L^2(z_{\ca T})}(\bfa f,\bfa g)=\Vert \bfa f_{i} - \bfa f_{j}\Vert_{
\bfa L^2(z_{\ca T})}\geq \frac{\epsilon}{2}\;\;\;\; \text{for all }1\leq i<j\leq m,
\end{equation*}
with  $\bfa\Gamma(\ca T)_{t,k}=\sqrt{N_k}\bfa \Gamma_{t,k}$ and $\bfa \Gamma$ is the mixing matrix of RGP. 
\end{lemma}

\begin{theorem}\label{thm:24} When the $\zt$ is sampled according to ergodic $\ezt$ with mixing matrix $\bfa \Gamma$, for all $\epsilon\in (0,1]$ we have:
\begin{align*}
   N(\pa H,d_{\bfa L^2(\zt)},\epsilon)&=\prod_{t=1}^T\bigg(\frac{32e\Vert\bfa\Gamma\Vert_2}{9T \epsilon^4}\bigg)^{2\vc t}\\
   \Vert\ecrad{\pa H}\Vert_1&=\ca O\bigg(\sqrt{\frac{1}{T|V|}\sum_{t=1}^T\vc i}\bigg)
\end{align*}
\label{146}
\end{theorem}
 
\begin{remark}
  Our result suggests that $\Vert\bfa \Gamma\Vert_2$ mainly influences the logarithmic factor in the bound. When $\bfa \Gamma=\bfa I$, the communication is fully eliminated and the mixing case reduced to the independent case.
\end{remark}


\section{Consistency Guarantees}
In this section, we leverage the newly proposed complexities to obtain uniform learnability, and furthermore, the consistency of the Empirical Risk Minimizer (ERM). The discussion focuses on the uniform ergodicity but is paired with the independent regime as a comparison.

\subsection{Empirical Risk Minimizer}
The empirical risk minimizer is a set of optimal hypotheses acting on the samples. For a given problem, we are interested in two main questions regarding the empirical risk minimizer. \textbf{(1)} How the empirical risk minimizer generalizes to unknown samples. \textbf{(2)} How the empirical risk minimizer compared with the Bayes risk minimizer. The first question can be understood via uniform learn-ability. The second question is more involved to answer, where the expected regret is a better characterization for online learning.

The empirical risk minimizer (ERM) $\bfa h_\erm\in\pa H$  for some $\zt$ drawn from $\bb Z$ according to $\ezt$ satisfies:
\begin{equation*}
   h_{\erm,t}=\argmin_{h\in\ca H_t}\sum_{j\in  [N_t]}\frac{1}{TN_t} L(h(X_{t,j}),Y_{t,j})\;\;\text{ for all }t\in[T]
\end{equation*}
An important utility of structural Rademacher complexity is upper-bounding the expected regret of ERM. The following bound gives the first moment control on the regret.
\begin{lemma}[Expected Regret Bound of ERM]\label{lm:26}
\label{1}Let $\bb Y=\{-1,1\}$ and $L(h(x),y)=\frac{1-h(x)y}{2}$, we have
$   \bb E_{\ezt} \ca R_{\ca Z_{\ca T}}(\bfa h_\erm)\leq\bb E_{\ezt}\Vert\crad{\pa H}\Vert_1
$\end{lemma}
 
Remarkably, if we consider the degenerated case where $|V|=1$, we can easily see that the expected 1-norm of structural Rademacher complexity is a strict lower bound on the sequential Rademacher complexity~\citep{rakhlin2015sequential}, hence we gave a better guarantee for the expected regret. 

The bound on expectation can also be strengthened to the non-asymptotics. However, this relies on more assumptions, which we discussed in the following sections.

\subsection{Non-communicative Networks}
Consider a non-communicative network, where the process is assumed to be independent. We can establish uniform learnability for it.

\begin{theorem}
\label{th:gen}
Let $Z_{\ca T}$ be a $\{-1,1\}$ labeled sample drawn from $\bb Z$ according to an independent RGP $\ca Z_{\ca T}$.
Assuming the loss function $L$ is given by $L(h(x),y)=\frac{1-yh(x)}{2}$.
Then, for any $\delta\in(0,1)$, with probability at least $1-\delta$ over $\zt$, the following holds for all $\bfa h\in\pa H$.
    \begin{align*}
        &R_{\ezt}(\bfa h)\leq \wh R_{\zt}(\bfa h)+\Vert\ecrad{\pa H}\Vert_1+\frac{3}{T}\sqrt{\sum_{t=1}^T\frac{1}{2N_t}\log\frac{2}{\delta}}
    \end{align*}
\end{theorem}
 
With this result, we can obtain the consistency of ERM for this problem.
\begin{theorem}[ERM on Non-communicative networks]\label{thm:ERMn}
The empirical risk minimizer for the non-communicative network problem yield the following guarantees on the expected regret with probablity at least $1-\delta$ for all $\delta\in(0,1)$:
\begin{align*}
   \ca R_{\ezt}(\bfa h_\erm)\leq 2\Vert\ecrad{\pa H}\Vert_1+\frac{6}{T}\sqrt{\sum_{t=1}^T\frac{1}{2N_t}\log\frac{2}{\delta}}
\end{align*}
\end{theorem}
\begin{remark}
Note that the non-communicative case has faster decay with regard to $T$ as compared to the number of activated vertices $N_t$ in each timestamp. If the selection of activated vertices in each timestamp is arbitrary, then there will be a trade-off between exploration (seeing more timestamps) and exploitation (seeing more samples in the same timestamp).
\end{remark}

\subsection{Uniform Ergodic Network}
Here we formally obtain the learning guarantees for a uniform ergodic network. With the aforementioned assumptions, we can also achieve similar guarantees for the ERM on this problem. We first present a simplified version of uniform learnability using the mixing matrix. Then we refined the global ergodic setting to gave sharper analysis, followed by the guarantees on the expected regret. 
\begin{theorem}[Ergodic RGP]
\label{ergodic}
Let $Z_{\ca T}$ be a $-1/1$ labeled sample drawn from $\Omega_1^T$ according to an uniform ergodic RGP $\ca Z_{\ca T}$. Let $L$ be the loss defined by $L(\wh y,y)=\frac{1-y\wh y}{2}$. Then for any $\delta>0$, with probability at least $1-\delta$, for all $\bfa h\in \pa H$ where $\pa H$ is $1$-Lipschitz, the following holds:
\begin{align*}
     R_\zt(\bfa h)&\leq\wh R_\zt(\bfa h)+\Vert\ecrad{\pa H}\Vert_1+\frac{3}{T}\Vert\bfa N^{\frac{1}{2}}\bfa\Gamma\cdot\bfa c\Vert_2\sqrt{2\log\frac{2}{\delta}}
\end{align*}
with $\bfa c=(\frac{1}{N_1},\ldots,\frac{1}{N_T})^\top$ and for all $t<k$:
{\begin{align*}
     &\Gamma_{t,k}=\begin{cases}
    k_0\rho^{k-t} & \text{if}\ k\geq t\\
     1,& \text{if}\ k= t\\
      0, & \text{otherwise}\end{cases}
      \end{align*}}
\end{theorem}
Intuitively, we found that the non-asymptotic rate is $O\big (\frac{1}{T\sqrt{|V|}}\big )$. Coupled with the upperbound on structural Rademacher complexity we can achieve the right upperbound of order $O\big(\frac{1}{\sqrt{T|V|}}\big)$. This guarantees the consistency. However, we may expect a faster concentration, done through a refinement argument.

Note that in the previous bound, we consider a batch of vertices as a whole. This can sometimes yield over-pessimistic estimation. We can refine the upper bound through the introduction of reachable sets.
\begin{definition}[Reachable Set]
We say a set $V_{t,j}^k\subset \ca V_k$ is reachable from the vertex indexed by $(t,j)$ if $\forall s\in V_{t,j}^k$, the message sent by $(t,j)$ reaches $s$. We denote $card( V_{t,j}^k)=\ca N_{t,j}^k$
\end{definition}
And the following bound gives a nicer guarantee.
\begin{theorem}[Refined upper bound for globally ergodic RGP]
\label{refine}
Under the analogous conditions and notations of theorem \ref{ergodic}, for any $\delta\geq 0$, with probability at least $1-\delta$:
\begin{align*}
         R_\ezt(\bfa h)&\leq\wh R_\zt(\bfa h)+\Vert\ecrad{\pa H}\Vert_1+3\sqrt{\sum_{t=1}^T\sum_{j=1}^{N_t}\bigg(\sum_{k=t+1}^T\frac{\ca N_{t,j}^k k_0\rho^{k-t}}{N_k^2T}\bigg)^2}\sqrt{2\log\frac{2}{\delta}}
\end{align*}
when the RGP is globally ergodic. 
\end{theorem}
\begin{remark}
  The refined bound is remarkable since it yields a sharper tail, which converges in the order $O\big (\frac{1}{|V|^{3/2}T})$ in sparse networks. When the network is dense (e.g., $\ca N_{t,j}^k=\Theta(|N_k|)$ ), the rate remains the same with the unrefined one.
\end{remark} 
  The following theorem then guarantees the consistency of ERM on learning the uniform ergodic network.
 
\begin{theorem}[ERM on Ergodic Network]\label{thm:34}
The empirical risk minimizer for the uniform ergodic network yield the following with probability at least $1-\delta$:
\begin{align*}
    \ca R_{\ezt}(\bfa h_\erm)\leq 2\bb E_\ezt\Vert\crad{\pa H}\Vert_1+2\sqrt{\sum_{t=1}^T\sum_{j=1}^{N_t}\bigg(\sum_{k=t+1}^T\frac{\ca N_{t,j}^k k_0\rho^{k-t}}{N_k^2T}\bigg)^2}\sqrt{2\log\frac{2}{\delta}}
\end{align*}
\end{theorem}

\section{Open Problems and Future Work}
We conclude with several open problems for future exploration.
\begin{compactenum}
\item The generalization bound for uniform ergodic RGP is improvable as the uniform argument can be relaxed.
\item The uniformity in the ergodic condition is very strong. We conjectured that weaker regularity conditions also yield uniform convergence.
\item We did not include efficient algorithms in this work and left those as future work.
\end{compactenum}
\section{Conclusions}
This work studied the problem of learning a network with communicative vertices. We studied the uniform ergodic network and compared it with the non-communicative counterpart. We introduced the structural Rademacher complexity and discussed its relationship with the VC theory. Our bound yielded a sharper bound on the first moment error of the stochastic processes than the previous work. We gave uniform learn-ability via the martingale method and Marton coupling, which leads to the consistency guarantee of ERM. 


\bibliography{bib}

\appendix
\section{Proof of Lemma \ref{contract}}
The proof idea follows that of $\epsilon$-maximizer argument. 
First we decouple the expectation and pick only a single random variable in the empirical complexity. We introduce the two imcomplete sum as:
\begin{align*}
    \ca U_1^{T-1}(h_1^{T}) &=\frac{1}{N_t}\sum_{t=1}^{T-1}\sum_{j\in \ca V_t}\sigma_{t,j}(\Phi_t\circ h_t)(X_{t,j})\\
    u_2^{N_T}(h)&=\frac{1}{N_t}\sum_{j=2}^{N_t}\sigma_{T,j}(\Phi_T\circ h)(X_{T,j})
\end{align*}
where we reindex $\ca V_t=\{1,\ldots,N_t\}$ for simplicity.
\begin{align*}
     &\bb E_{\bfa \sigma}\frac{1}{T}\sup_{\bfa h\in\pa H}\sum_{t=1}^T\frac{1}{N_t}\sum_{j\in V_t} \sigma_{t,j}(\Phi_t\circ h_t)(X_{t,j})\\
     &=\frac{1}{T}\bb E_{\bfa\sigma_1^{T-1}}\bigg[\bb  E_{\sigma_{T,2},...,\sigma_{T,N_t}}\bigg[\bb E_{\sigma_{T,1}}\bigg[\sup_{\bfa h\in\pa H}\bigg(\ca U_1^{T-1}(h_1^{T})+u_2^{N_T}(h_T)+\frac{\sigma_{T,1}}{N_t}(\Phi_T\circ h_T)(X_{T,1})\bigg)\bigg]\bigg]\bigg]
\end{align*}
Then we introduced the $\epsilon$-maximizer. Using the property of supremum, for any $\epsilon>0$, there exist $\fk h_1,\fk h_2\in\pa H$ s.t.
\begin{align*}
    \ca U_1^{T-1}(\fk h_1)+u_2^{N_T}(\fk h_{1,T})+\frac{1}{N_T}(\Phi_T\circ \fk h_{1,T})(X_{T,1})&\geq (1-\epsilon)\bigg[\sup_{\bfa h\in\pa H}\ca U_1^{T-1}(h_1^T)+u_2^{N_T}(h_T)+\frac{1}{N_T}(\Phi_T\circ h_T)(X_{T,1})\bigg]\\
    \ca U_1^{T-1}(\fk h_2)+u_2^{N_T}(\fk h_{2,T})-\frac{1}{N_T}(\Phi_T\circ \fk h_{2,T})(X_{T,1})&\geq (1-\epsilon)\bigg[\sup_{\bfa h\in\pa H}\ca U_1^{T-1}(h_1^T)+u_2^{N_T}(h_T)-\frac{1}{N_T}(\Phi_T\circ h_T)(X_{T,1})\bigg]
\end{align*}
Thus, for any $\epsilon>0$, we have
\begin{align*}
     &(1-\epsilon)\bb E_{\sigma_{T,1}}\bigg[\sup_{\bfa h\in\pa H}\ca U_1^{T-1}(h_1^T)+u_2^{N_T}(h_T)+\frac{\sigma_{T,1}}{N_T}(\Phi_T\circ h_T)(X_{T,1})\bigg]\\
     &=(1-\epsilon)\bb E_{\sigma_{T,1}}\frac{1}{2}\bigg[\sup_{\bfa h\in\pa H}\ca U_1^{T-1}(h_1^T)+u_2^{N_T}(h_T)-\frac{1}{N_T}(\Phi_T\circ h_T)(X_{T,1})\bigg]\\
     &+(1-\epsilon)\bb E_{\sigma_{T,1}}\frac{1}{2}\bigg[\sup_{\bfa h\in\pa H}\ca U_1^{T-1}(h_1^T)+u_2^{N_T}(h_T)-\frac{1}{N_T}(\Phi_T\circ h_T)(X_{T,1})\bigg]\\
     &\leq \frac{1}{2}\bigg(\ca U_1^{T-1}(\fk h_1)+u_2^{N_T}(\fk h_{1,T})+\frac{1}{N_T}(\Phi_T\circ \fk h_{1,T})(X_{T,1})\bigg)\\
     &+\frac{1}{2}\bigg(\ca U_1^{T-1}(\fk h_2)+u_2^{N_T}(\fk h_{2,T})-\frac{1}{N_T}(\Phi_T\circ \fk h_{2,T})(X_{T,1})\bigg)\\
     &\leq\frac{1}{2}\bigg(\ca U_1^{T-1}(\fk h_1)+u_2^{N_T}(\fk h_{1,T})+\ca U_1^{T-1}(\fk h_2)+u_2^{N_T}(\fk h_{2,T})+\frac{l}{N_t}|\fk h_{1,T}(X_{T,1})-\fk h_{2,T}(X_{T,1})|\bigg)\\
     &\leq \frac{1}{2}\sup_{\bfa h\in\pa H}\bigg(\ca U_1^{T-1}(h_1^{T})+u_2^{N_T}(h_T) +\frac{l}{N_T}h_T(X_{T,1})\bigg)+\frac{1}{2}\sup_{\bfa h\in\pa H}\bigg(\ca U_1^{T-1}(h_1^T)+u_2^{N_T}(h_T) -\frac{l}{N_T}h_T(X_{T,1})\bigg)\\
     &=\bb E_{\sigma_{T,1}}\bigg[\sup_{\bfa h\in\pa H}\ca U_1^{T-1}(h_1^T)+u_2^{N_T}(h_T) +\frac{\sigma_{T,1}l}{N_T}h(X_{T,1})\bigg]
\end{align*}
When $\epsilon\rightarrow 0$, we have
\begin{align*}
    &\lim_{\epsilon\rightarrow0}(1-\epsilon)\bb E_{\sigma_{T,1}}\bigg[\sup_{\bfa h\in\pa H}\ca U_1^{T-1}(h_1^T)+u_2^{N_T}(h_T)+\frac{\sigma_{1,1}}{N_T}(\Phi_T\circ h_T)(X_{T,1})\bigg]\\
    &=\bb E_{\sigma_{T,1}}\bigg[\sup_{\bfa h\in\pa H}\ca U_1^{T-1}(h_1^T)+u_2^{N_T}(h_T)+\frac{\sigma_{1,1}}{N_T}(\Phi_T\circ h_T)(X_{T,1})\bigg]\\
    &\leq \bb E_{\sigma_{T,1}}\bigg[\ca U_1^{T-1}(h_1^T)+u_2^{N_T}(h_T) +\frac{\sigma_{1,1}l}{N_t}h_T(X_{T,1})\bigg]
\end{align*}
Proceeding the same way for $\{\sigma_{T,2},...,\sigma_{T,N_t}\}$, we have
\begin{align*}
    &\bb E_{\bfa\sigma_{T}}\bigg[\sup_{\bfa h\in\pa H}\ca U_1^{T-1}(h_1^T)+\frac{1}{N_T}\sum_{j\in [N_T]}\sigma_{T,j}(\Phi_T\circ h_T)(X_{T,j})\bigg]\\
    &\leq\bb E_{\bfa\sigma_{T}}\bigg[\sup_{\bfa h\in\pa H}\ca U_1^{T-1}(h_1^T)+\frac{1}{N_T}\sum_{j\in [ N_T]}\sigma_{T,j}lh(X_{T,j})\bigg]
\end{align*}
Then proceeding the same way from $T-1$ to $1$ completes the proof.

\section{Proof of Lemma \ref{lm:7}}
We use $\bfa\sigma_t=(\sigma_{t,1},...,\sigma_{t,N_t})$ to denote the random vector constructed by $N_t$ i.i.d. Rademacher variables, hence
\begin{align*}
    \ecrad{\pa H}_t=\bb E_{\bfa\sigma_t}\sup_{h\in\ca H_t}\sum_{j\in\ca V_t}\frac{\sigma_{t,j}h(X_{t,j})}{TN_t}
\end{align*}
And we introduce the random matrix whose rows being the feature vector of activated vertices at timestamp $t$: $\bfa X^\prime_{t}=( X^\prime_{t,\pi_t(1)},\ldots X^\prime_{t,\pi_t(N_t)})$ such that $X^\prime_{t,\pi_t(j)}= X_{t,\pi_t(j)}$ for all $j\in[N_t]\setminus\{j_0\}$ for some arbitrary $j_0$.
\begin{align*}
    \bb E_{\bfa\sigma_t}\sup_{h\in\ca H_t}\sum_{j\in\ca V_t}\frac{\sigma_{t,j}h(X_{t,j})}{TN_t}-\bb E_{\bfa\sigma_t}\sup_{h\in\ca H_t}\sum_{j\in\ca V_t}\frac{\sigma_{t,j}h(X^\prime_{t,j})}{TN_t}\leq\frac{M}{TN_t}
\end{align*}
Applying Mcdarmid's inequality and using the conditional independence assumption, we have
\begin{align*}
    \bfa P\bigg(\ecrad{\pa H}_t-\crad{\pa H}_t\geq\epsilon\bigg | \pa Z_1^{t-1}\bigg)\leq\exp\bigg(\frac{-2\epsilon^2}{\sum_{t=1}^{N_t}(\frac{M}{TN_t})^2}\bigg)
\end{align*}
by substituting the R.H.S. with $\delta$, we complete the proof of the first inequality.

To prove the second inequality, we note that  with probability at least $1-\frac{\delta}{T}$, the following holds for all $t\in[T]$:
\begin{align*}
     \crad{\pa H}_t\leq\ecrad{\pa H}_t +\frac{M}{T}\sqrt{\frac{1}{2N_t}\log\frac{T}{\delta}}
\end{align*}
Using the union bound, we have with probability at least $1-\delta$, for all $t\in[T]$, the following holds:
\begin{align*}
     \crad{\pa H}\leq\ecrad{\pa H} +\frac{M}{T}\sqrt{\log\frac{T}{\delta}}\sum_{t=1}^T\frac{1}{\sqrt{2N_t}}
\end{align*}
Summing them up and we complete the proof of the second inequality.

For the third inequality, when the independence condition holds,  we note that:
\begin{equation*}
    \bb E_{\bfa\sigma}\sum_{t=1}^T\sum_{j\in\ca V_t}\frac{\sigma_{t,j}h(X_{t,j})}{TN_t}-\bb E_{\bfa\sigma}\sum_{t=1}^T\sum_{j\in\ca V_t}\frac{\sigma_{t,j}h(X^\prime_{t,j})}{TN_t}\leq\frac{M}{TN_t}
\end{equation*}
when $\pa X_j=\pa X_j^\prime$ for all $j\in[T]\wedge j\neq t$
Hence applying Mcdarmid's inequality to $\Vert\ecrad{\pa H}\Vert_1$ yield that:
\begin{align*}
 \bfa P(\ecrad{\pa H}-\crad{\pa H}\geq\epsilon)\leq\exp\bigg(\frac{2\epsilon^2}{\sum_{t=1}^T\sum_{j\in\ca V_t}(\frac{M}{TN_t})^2}\bigg)
\end{align*}
substituting the R.H.S. with $\delta$ and we complete the proof.

\section{Proof of Theorem \ref{chaining}}
First we review the function in Orlicz space, defined by:
\begin{definition}[Orlicz function]
A function $\psi$ is called Orlicz function if $\psi$ is convex, increasing and satisfies:
\begin{align*}
    \psi(0)=0.\;\;\;\psi(x)\rightarrow\infty\;\text{ as }x\rightarrow\infty
\end{align*}
\end{definition}
For a given Orlicz function, the Orlicz norm of a random variable is defined by
\begin{align*}
    \Vert X\Vert_{\psi}:=\inf\{t>0: \bb E\psi(|X|/t)\leq 1\}
\end{align*}
Any sub-gaussian random variables $X$ will have bounded Orlicz norm with $\psi_2=\exp(x^2)-1$ being the Orlicz function.
\begin{align*}
    \Vert X\Vert_{\psi_2}\leq C
\end{align*}
The Orlicz norm admits the following useful property
\begin{lemma}
Every sub-gaussian random variable $X$ satisfies the following bounds:
\begin{align*}
    \bb P(|X|\geq t)&\leq 2\exp\bigg (\frac{-ct^2}{\Vert X\Vert_{\psi_2}^2}\bigg )\;\;\text{ for all }t\geq 0;\\
    \text{ if }\bb E X=0\;\text{ then }\bb E\exp(\lambda X)&\leq \exp(C\lambda^2\Vert X\Vert_{\psi_2}^2)\;\;\text{ for all }\lambda\in\bb R.
\end{align*}
for some universal constant $c$ and $C$.
\end{lemma}
\begin{lemma}
\label{lm:66}
Let $X_1,\ldots X_N$ be independent, mean zero, sub-gaussian random variables. Then $\sum_{i=1}^N X_i$ is also a sub-gaussian random variable, and
\begin{align*}
    \bigg\Vert\sum_{i=1}^NX_i\bigg\Vert^2_{\psi_2}\leq C\sum_{i=1}^N\Vert X_i\Vert^2_{\psi_2}
\end{align*}
where $C$ is a universal constant.
\end{lemma}
\begin{lemma}\label{lm:777}
Let $X_1,\ldots, X_N$ be a sequence of sub-gaussian random variables not necessarily independent, then the following holds
\begin{align*}
    \bb E\sup_{i\leq N}|X_i|\leq CK\sqrt{\log N}
\end{align*}
where $K=\sup_{i\in[N]}\Vert X\Vert_{\psi_2}$
\end{lemma}
Fixing the random graph process and we consider the following process
\begin{align*}
    \pa G(\bfa h)&:=\sum_{t=1}^T\frac{1}{TN_t}\sum_{i=1}^{N_t}\epsilon_{t,i}h_t(X_{t,i})
\end{align*}
First we found that for the Rademacher random variable $\epsilon$, the following holds:
\begin{align*}
    \Vert \epsilon \Vert_{\psi_2}=\frac{1}{\sqrt{\ln 2}}
\end{align*}
Using lemma \ref{lm:66}, we observe that:
\begin{align*}
    \Vert\pa G(\bfa h)-\pa G(\bfa g)\Vert_{\psi_2}^2&\leq C\sum_{t=1}^T\sum_{i=1}^{N_t}\frac{(h_t(X_{t,i})-g_t(X_{t,i}))^2}{T^2N_t^2}\frac{1}{\ln 2}\\
    &\leq C_0\frac{1}{T}\frac{1}{\inf_t N_t}\bigg(\sum_{t=1}^T\sum_{i=1}^{N_t}\frac{(h_t(X_{t,i})-g_t(X_{t,i}))^2}{TN_t}\bigg)
\end{align*}
which implies that
\begin{align*}
    \Vert\pa G(\bfa h)-\pa G(\bfa g)\Vert_{\psi_2}&\leq C\sqrt{\sum_{t=1}^T\sum_{i=1}^{N_t}\frac{1}{T^2N_t^2}(h_t(X_{t,i})-g_t(X_{t,i}))^2}\\
    &\leq C\Vert \bfa h- \bfa g\Vert_{L^2(\zt)}\cdot\sqrt{\frac{1}{T\inf_tN_t}}
\end{align*}
Using lemma \ref{lm:777}, we see that for some finite set $\pa A\subset\pa H$ and a sample $\zt$, we have
\begin{align*}
    \bb E_{\bfa\sigma}\sup_{\bfa h\in \pa A}\pa G(\bfa h)&\leq C_1\Vert\pa G(\bfa h)\Vert_{\psi_2}\sqrt{\log |\pa A|}\\
    &\leq C_2\Vert \bfa h\Vert_{L^2(\zt)}\sqrt{\log|\pa A|}
\end{align*}
for some absolute constant $C_2$

Therefore, we can find a sequence of subsets $\ca A=(\pa A_0,...,\pa A_{n})$ with $\pa A_k$ being the $2^{-k}diam(\pa H)$-net for $\pa H$. By the definition of $\epsilon$-net, for all $\bfa h\in\pa H$, there exists $\pi^k(\bfa h)=(\pi^k_1(\bfa h),...,\pi^k_T(\bfa h))^\top\in \pa A_k$, such that $d_{\bfa L^2(\zt)}(\bfa h,\pi^k(\bfa h))\leq 2^{-k}diam(\pa H)$. By the duality between the 1-norm and 2-norm, we see that
\begin{align*}
    |\pa G(\bfa f)-\pa G(\bfa g)| &=\sum_{t=1}^T\sum_{i\in[N_t]}\frac{1}{N_t T}(f_t(X_{t,i})-g_t(X_{t,i}))\\
    &\leq\bigg(\sum_{t=1}^T\sum_{i=1}^{N_t}\frac{1}{N_t^2T^2}(f_t(X_{t,i})-g_t(X_{t,i}))^2\bigg)^{\frac{1}{2}}\bigg(\sum_{i=1}^TN_t\bigg)^{\frac{1}{2}}\\
    &\leq \bigg(\sum_{t=1}^T\sum_{i=1}^{N_t}\frac{1}{N_tT}(f_t(X_{t,i})-g_t(X_{t,i}))^2\bigg)^{\frac{1}{2}}\bigg(\sum_{i=1}^TN_t\bigg)^{\frac{1}{2}}\sqrt{\frac{1}{T\inf_tN_t}}
\end{align*}
In particular, we let $\pa A_0=\{\bfa h_0\}$ such that $\bfa h_0=\argmin_{\bfa h}\sup_{\bfa h_1\in\pa H}\Vert\bfa h_0-\bfa h\Vert_{L^2(\bfa \mu)}$ 
\begin{align*}
    \Vert\ecrad{\pa H}\Vert_1&=\bb E_{\bfa\sigma}\sup_{\bfa h\in\pa H}\pa G(\bfa h)\\&=\bb E_{\bfa\sigma}\sup_{\bfa h\in\pa H}\bigg [(\pa G(\bfa h)-\pa G(\pi^n(\bfa h)))+\sum_{k=1}^{n-1}\big (\pa G(\pi^{k}(\bfa h))-\pa G(\pi^{k-1}(\bfa h))\big)+\pa G(\pi^0(\bfa h))\bigg ]\\
    &\leq\bb E_{\bfa\sigma}\sup_{\bfa h\in\pa H}\bigg [(\pa G(\bfa h)-\pa G(\pi^n(\bfa h)))\bigg]+\sum_{k=1}^{n-1}\bb E_{\bfa\sigma}\sup_{\bfa h\in\pa H}\bigg[\pa G(\pi^{k}(\bfa h))-\pa G(\pi^{k-1}(\bfa h)) \bigg ]\\
    &+\bb E_{\bfa\sigma}\sup_{\bfa h\in\pa H} \pa G(\pi^0(\bfa h))\\
    &\leq 2^{-N}diam(\pa H)\bigg(\sum_{i=1}^TN_t\bigg)^{\frac{1}{2}}\sqrt{\frac{1}{T\inf_tN_t}}\\
    &+C\sum_{k=1}^{n-1}\Vert\pa G(\pi^{k}(\bfa h))-\pa G(\pi^{k-1}(\bfa h))\Vert_{\psi_2} \sqrt{\log|\pa A_{k}|\cdot |\pa A_{k-1}|}\\
    &\leq 2^{-N}diam(\pa H)\bigg(\sum_{i=1}^TN_t\bigg)^{\frac{1}{2}}\sqrt{\frac{1}{T\inf_tN_t}}\\
    &+ C\sqrt{\frac{1}{T\inf_tN_t}}\sum_{k=1}^{n-1}d_{\bfa L^2(\bfa\zt)}(\pi^k(\bfa h)-\pi^{k-1}(\bfa h))\sqrt{\log |\pa A_k|\cdot|\pa A_{k-1}|}
\end{align*}

Note that we can choose $n$ large enough such that $|\pa A_n|=|\pa H|$ and this result can be extended to countable set. Moreover, note that the maximum in the $k$-th term contains at most $|\pa A_k||\pa A_{k-1}|\leq |\pa A_k|^2$ terms,
\begin{align*}
    &d\lzt{2}(\pi^k(\bfa h),\pi^{k-1}(\bfa h))\\
    &\leq d\lzt{2}(\bfa h,\pi^{k-1}(\bfa h))+d\lzt{2}(\bfa h,\pi^k(\bfa h))\\
    &\leq 3\times 2^{-k}diam(\pa H)
\end{align*}
Hence
\begin{align*}
    &\bb E_{\bfa\sigma}\bigg[\sup_{\bfa h\in\pa H}\sum_{t=1}^T\frac{1}{TN_t}\sum_{i\in[N_t]}\sigma_{t,i}(\pi^k_t(\bfa h)(x_{t,j})-\pi_{t}^{k-1}(\bfa h)(x_{t,j}))\bigg]\\
    &\leq C\sqrt{\frac{1}{T\inf_tN_t}}\cdot 2^{-k}diam(\pa H)\sqrt{\log N(\pa H,d\lzt{2},2^{-k}diam(\pa H))}
\end{align*}
And without loss of generality, we can always set $\bfa 0=|\pa A_0|$ and $\ecrad{\pa A_0}=0$. We then immediately have, using the fact that $N(\pa H,d\lzt{1},2^{-k})\leq N(\pa H,d\lzt{2},2^{-k})$
\begin{align*}
    \Vert\ecrad{\pa H}\Vert_1&\leq C\sqrt{\frac{1}{T\inf_tN_t}}\sum_{k= 1}^n2^{-k}diam(\pa H)\sqrt{\log N(\pa H,d\lzt{2},2^{-k}diam(\pa H))}\\
    &\leq C\sqrt{\frac{1}{T\inf_tN_t}}\sum_{k=1}^\infty\int_{2^{-k}diam(\pa H)}^{2^{-k+1}diam(\pa H)}\sqrt{\log N(\pa H,d\lzt{2},\epsilon)}d\epsilon\\
    &\leq C\sqrt{\frac{1}{T\inf_tN_t}}\int_{0}^{diam(\pa H)}\sqrt{\log N(\pa H,d\lzt{2},\epsilon)}d\epsilon\\
    &\leq C\sqrt{\frac{1}{T\inf_tN_t}}\int_{0}^{\infty}\sqrt{\log N(\pa H,d\lzt{2},\epsilon)}d\epsilon\\
\end{align*}
The infinite case can be approximated with separability assumption, which we omitted here and refer the interested reader to ~\citet{boucheron2013concentration}.

\section{Proof of Lemma \ref{lm:13}}
 Note that for $\bfa f,\bfa g\in\pa H$, we have
 \begin{equation*}
     d_{\bfa L^2(\zt)}(\bfa f,\bfa g)\leq d_{\bfa L^\infty(\zt)}(\bfa f,\bfa g)\leq d_{\infty}(\bfa f,\bfa g)
 \end{equation*}
 and an $\epsilon$-net of matrix $d_{\bfa L^2(\zt)}(\bfa f,\bfa g)$ is also an $\epsilon$-net of $ d_{\bfa L^\infty(\zt)}$, which completes the proof.

\section{Proof of Lemma \ref{lm:16}}
Note that $\mtt{vc}(\pa C)=\sum_{i=1}^T\mtt{vc}(\ca C_i)$ holds in the independent case from the definition. 

An example of non i.i.d. $\zt$ is the random process defined by $\pa X_{t+1}=\pa X_{t}$ with concept classes $\pa C=(\ca C_1,...,\ca C_T)$ such that $\ca C_i\subset\ca C_{i+1}$ for all $i\in[T-1]$. In this problem we have $\mtt{vc}(\pa C)=T \mtt{vc}(\ca C_1)<\sum_{i=1}^T\mtt{vc}(\ca C_i)$

\section{Proof of Lemma \ref{lm:18}}
Note that when $\epsilon\leq 1$, assuming $N=\{f_1,...,f_m\}$ is an $\epsilon$-covering for $(\pa C, d_\infty)$, then for all $C\in\ca C$, there exists $i\in[m]$ such that
\begin{equation*}
    d_\infty(f_i,\bfa C)\leq\epsilon\leq 1
\end{equation*}
And we have:
\begin{equation*}
    d_{\infty}(\bfa C,\bfa C^\prime)=1\;\;\;\; \text{ for all }\bfa C\neq \bfa C^\prime\text{ and } \bfa C,\bfa C^\prime\in \pa C
\end{equation*}
It implies that:
\begin{equation*}
    N(\pa C,d_{\infty},\epsilon)=|\pa C|\;\;\text{for all }\epsilon\leq1
\end{equation*}
Using similar argument, we have, $N(\pa H,d_{L^\infty( \zt)},\epsilon)=(\pa C,d_{L^\infty( \zt)},\epsilon) = \Pi_{\pa H}(Z_{\ca T})$ for all $\epsilon\leq 1$

\section{Proof of Lemma \ref{lm:19}}
The proof of Sauer's lemma relies on ~\citet{pajor1985sous}, which is stated as follows
\begin{lemma}
\label{lm90}
For any class $\ca C$ of subsets of $\bb X$, 
$    |\ca C|\leq |\{ \bfa I\subseteq \bb X:\bfa I\text{ is shattered by }\ca C\}|
$
\end{lemma}
We can then move on to extend Sauer's lemma to bound the growth function for the RGP:
From lemma \ref{lm90}, for any class $\pa C\subset \bb X^T$, we have:
\begin{align*}
    |\pa C\cup \zt|&\leq \prod_{t=1}^T|\{I_t\subseteq \ca V_t: I_t\text{ is shattered by }\ca C_t\}|\\
    &\leq\prod_{t=1}^T|\{ I_t\subseteq \ca V_t: |I_t|\leq\min\{\mtt{vc}(\ca C_t), N_t\}\}|=\prod_{t=1}^T\sum_{k=0}^{\min\{\mtt{vc}(\ca C_t), N_t\}}\binom{N_t}{k}
\end{align*}
Then we use the binomial theorem:
\begin{equation*}
    \sum_{k=0}^d\binom{n}{k}\leq\bigg(\frac{en}{d}\bigg)^d 
\end{equation*}
to complete the proof.

\section{Proof of Lemma \ref{lm:21}}
Note that for any $z_{\ca T}$ and $z_{\ca T}^\prime$ that only differ in $z_{t,j}$, we have
\begin{equation*}
    \Vert \bfa f_i-\bfa f_j\Vert_{\bfa L^2(z_{\ca T})}^2-\Vert \bfa f_i-\bfa f_j\Vert_{\bfa L^2(z^\prime_{\ca T})}^2\leq\frac{1}{N_tT}
\end{equation*}Using Mcdiarmid's inequality, for any $i\neq j$ and $\epsilon\leq \frac{1}{\sup_t N_tT}$ we have
\begin{align*}
    \bfa P\bigg[\Vert \bfa f_i-\bfa f_j\Vert_{\bfa L^2(z_{\ca T})}^2\leq\frac{\epsilon^2}{4}\bigg]&\leq\bfa P\bigg[\Vert \bfa f_i-\bfa f_j\Vert^2_{\bfa L^2(z_{\ca T})}-\Vert \bfa f_i-\bfa f_j\Vert^2_{\bb E\bfa L^2( Z_{\ca T_0})}\leq-\frac{3\epsilon^2}{4}\bigg]\\
    &\leq\exp\bigg(-\frac{9\epsilon^4T^2}{8\sum_{t=1}^T\frac{1}{N_t}}\bigg)
\end{align*}
Then the union bound yields that:
\begin{equation*}
    \bfa P\bigg[\Vert \bfa f_i-\bfa f_j\Vert_{\bfa L^2(z_{\ca T})}> \frac{\epsilon}{2}\text{ for all } i\neq j\bigg]\geq 1-\frac{m(m-1)}{2}\exp\bigg(-\frac{9\epsilon^4T^2}{8\sum_{t=1}^T\frac{1}{N_t}}\bigg)>0
\end{equation*}
 Then for all $\ca T$ with \begin{align*}\frac{T^2}{\sum_{t=1}^T\frac{1}{N_t}}>\frac{16}{9}\epsilon^{-4}\log m
 \end{align*}
 we can find an $z_{\ca T}$ such that $(f_1,...,f_m)$ is the  $\frac{\epsilon}{2}$-packing of $\pa H$ on the $z_{\ca T}$.
\section{Proof of Theorem \ref{thm:22}}
Let $\{\bfa f_1,...,\bfa f_m\}$ be a maximal $\epsilon$-packing of $(\pa H,d_{\bb E\bfa L^2(\zt)})$. Due to Lemma \ref{lm121}, there exists $z_{\ca T}$ with $\frac{T^2}{\sum_{t=1}^T\frac{1}{N_t}}>\frac{16}{9}\epsilon^{-4}\log m$ such that $\{\bfa f_1,...,\bfa f_m\}$ is still a $\frac{\epsilon}{2}$-packing of $(\ca C,d_{L^2(Z_{\ca T})})$. Hence, we let \begin{align*}N_t=\frac{16}{9T^2}\epsilon^{-4}\log m=c\frac{\epsilon^{-4}}{T^2}\log m\end{align*} for all $t$ and use Sauer's lemma to obtain that
\begin{equation*}
  m\leq\Pi_{\pa H}(z_{\ca T})\leq\prod_{t=1}^T\bigg(\frac{eN_t}{\text{vc}(\ca H_t)}\bigg)^{\text{vc}(\ca H_t)}<\prod_{t=1}^T\bigg(\frac{16 e\log m}{9T^2\epsilon^4\mtt{vc}(\ca H_t)}\bigg)^{\vc t}
\end{equation*}
where the first inequality comes from that $d_{\bfa L^\infty(\zt)}(\bfa f,\bfa g)\geq d_{\bb E \bfa L^2(\zt)}(\bfa f,\bfa g)$ for all $\bfa f,\bfa g\in\pa H$.
Note that $\alpha\log x\leq x^\alpha$ and we obtain the following through algebraic manipulations
\begin{equation*}
   \log N(\pa H,d_{\bb E\bfa L^2(\ezt)},\epsilon)\leq m\leq\prod_{t=1}^T\bigg(\frac{32e}{9T\epsilon^4{\vc t}}\bigg)^{2\vc t}
\end{equation*}
We see that the above inequality is measure free, which indicates that for any empirical measure defined by $\prod_{t=1}^T\prod_{j=1}^{N_t}\delta_{Z_{t,j}=z_{t,j}}$, the inequality still holds. We then complete the proof.

\section{Proof of Lemma \ref{lm:23}}
We adopt similar martingale strategy as in the proof of theorem \ref{ergodic} the following inequality holds via some algebraic manipulations.
\begin{align*}
    \bfa P\bigg[\Vert \bfa f_i-\bfa f_j\Vert_{\bfa L^2(z_{\ca T})}^2\leq\frac{\epsilon^2}{4}\bigg]&\leq\bfa P\bigg[\Vert \bfa f_i-\bfa f_j\Vert^2_{\bfa L^2(z_{\ca T})}-\Vert \bfa f_i-\bfa f_j\Vert^2_{\bb E\bfa L^2( Z_{\ca T_0})}\leq-\frac{3\epsilon^2}{4}\bigg]\\
    &\leq\exp\bigg(-\frac{9\epsilon^4T^2}{8\Vert\bfa \Gamma(\ca T)\cdot\bfa n\Vert_2}\bigg)\\
    &\leq\exp\bigg(-\frac{9\epsilon^4T^2}{8\Vert\bfa \Gamma(\ca T)\Vert_2\Vert\bfa n\Vert_2}\bigg)
\end{align*}
with $\bfa n=(\frac{1}{N_1},...,\frac{1}{N_T})^\top$ and $\bfa \Gamma(\ca T)$ with similar notation as theorem \ref{ergodic} then the union bound yields that:
\begin{equation*}
    \bfa P\bigg[\Vert \bfa f_i-\bfa f_j\Vert_{\bfa L^2(z_{\ca T})}> \frac{\epsilon}{2}\text{ for all } i\neq j\bigg]\geq 1-\frac{m(m-1)}{2}\exp\bigg(-\frac{9\epsilon^4T^2}{8\Vert\bfa \Gamma(\ca T)\Vert_2\Vert\bfa n\Vert_2}\bigg)>0
\end{equation*}
 Then for some $\ca T$ with \begin{align*}
     \frac{1}{\sqrt{\sum_{t=1}^T\frac{1}{N_t^2}}}\geq\frac{8\Vert\bfa\Gamma(\ca T)\Vert_2}{9}T^{-2}\epsilon^{-4}\log m
 \end{align*} we can find an $z_{\ca T}$ such that $(f_1,...,f_m)$ is the  $\frac{\epsilon}{2}$-packing of $\pa H$ on the metric defined on $z_{\ca T}$, 

\section{Proof of Theorem \ref{thm:24}}
Let $\{f_1,...,f_m\}$ be a maximal $\epsilon$-packing of $(\pa H,d_{\bb E\bfa L^2(\zt)})$. Due to Lemma \ref{lm121}, there exists a $z_\ca T$ with $\frac{1}{\sum_{t=1}^T\frac{1}{N_t^2}}\geq\frac{16\Vert\bfa\Gamma(\ca T)\Vert_2}{9}T^{-2}\epsilon^{-4}\log\frac{m(m-1)}{2}$ such that $\{f_1,...,f_m\}$ is still a $\frac{\epsilon}{2}$-packing of $(\ca C,d_{L^2(Z_{\ca T})})$. Hence, we let $N_t=\frac{16\Vert\bfa\Gamma\Vert_2}{9}T^{-2}\epsilon^{-4}\log m$ and use Sauer's lemma to obtain that
\begin{equation*}
  m\leq\Pi_{\pa H}(z_{\ca T})\leq\prod_{t=1}^T\bigg(\frac{eN_t}{\text{vc}(\ca C_t)}\bigg)^{\text{vc}(\ca C_t)}<\prod_{t=1}^T\bigg(\frac{16e\Vert\bfa\Gamma\Vert_2\log m}{9T^2\vc{t}\epsilon^4}\bigg)^{\vc t}
\end{equation*}
where the first inequality comes from that $d_{\bfa L^\infty(\zt)}(\bfa f,\bfa g)\geq d_{\bb E \bfa L^2(\zt)}(\bfa f,\bfa g)$ for all $\bfa f,\bfa g\in\pa H$.
Note that $\alpha\log x\leq x^\alpha$ and we obtain the following through algebraic manipulations
\begin{equation*}
   N(\pa H,d_{\bb E\bfa L^2(\ezt)},\epsilon)\leq m\leq\prod_{t=1}^T\bigg(\frac{32e\Vert\bfa\Gamma\Vert_2}{9T\epsilon^4}\bigg)^{2\vc t}
\end{equation*}
We see that the above inequality is measure free, which indicates that for any empirical measure defined by $\prod_{t=1}^T\prod_{j=1}^{N_t}\delta_{Z_{t,j}=z_{t,j}}$, the inequality still holds. We then complete the proof.

\section{Proof of Lemma \ref{lm:26}}
By the definition of empirical risk minimizer we have
\begin{equation*}
     \wh R_{\zt}(\bfa h_\erm)=\inf_{\bfa h\in\pa H} \wh R_{\zt}(\bfa h)
\end{equation*}
Let 
\begin{align*}
    \bfa h^*=\argmin_{\bfa h\in\pa H}R_{\ezt}(\bfa h)
\end{align*}
And we have that:
\begin{align*}
    \ca R_{\ezt}(\bfa h_\erm)&=\bb E_{\ezt}R_{\ezt}(\bfa h_\erm)- \inf_{\bfa h\in\pa H}R_{\ca Z_{\ca T}}(\bfa h)\\
    &=\bb E_{\ezt}R_{\ca Z_{\ca T}}(\bfa h_\erm)-\bb E_{\ezt}\wh R_{ Z_{\ca T}}(\bfa h_\erm)+\bb E_\ezt \wh R_{Z_{\ca T}}(\bfa h_\erm)-\bb E_\ezt \wh R_\zt(\bfa h^*)\\
    &\leq\bb E_{\ezt}\sup_{\bfa h\in\pa H}\bigg[R_{\ca Z_{\ca T}}(h)-\wh R_{Z_{\ca T}}(h)\bigg]+\bb E_{\ezt}\inf_{h\in\pa H}\wh R_{\zt}(\bfa h)-\inf_{\bfa h\in\pa H}R_{\ezt}(\bfa h)\\
    &\leq\bb E_{\ezt}\sup_{\bfa h\in\pa H}\bigg[R_{\ca Z_{\ca T}}(h)-\wh R_{Z_{\ca T}}(h)\bigg]\\
    &=\bb E_{\ezt}\bigg[\sum_{t=1}^T\sup_{h_t\in\ca H_t}\bigg(\bb E_{\ezt}\bigg[\sum_{j=1}^N\frac{1}{TN_t}L(h_t(X_{t,j}),Y_{t,j})\bigg|\pa Z_1^{t-1}\bigg]-\sum_{j\in\ca V_t}\frac{1}{TN_t}L(h_t(X_{t,j}),Y_{t,j})\bigg)\bigg]\\
    &=\bb E_{\ezt}\bigg[\sum_{t=1}^T\bb E_{\ezt}\bigg[\sup_{h_t\in\ca H_t}\bigg(\bb E_{\ezt}\bigg[\sum_{j=1}^N\frac{1}{TN_t}L(h_t(X_{t,j}),Y_{t,j})\bigg|\pa Z_1^{t-1}\bigg]\\
    &-\sum_{j\in\ca V_t}\frac{1}{TN_t}L(h_t(X_{t,j}),Y_{t,j})\bigg)\bigg|\pa Z_1^{t-1}\bigg]\bigg]\\
    &\leq\bb E_{\ezt}\bigg[\sum_{t=1}^T\bb E_{\ezt}\bigg[\sup_{h_t\in\ca H_t}\bigg(\sum_{j\in\ca V_t}\frac{1}{N_tT}(L(h_t(X_{t,j}),Y_{t,j})-L(h_t(X^\prime_{t,j}),Y^\prime_{t,j}))\bigg)\bigg|\pa Z_1^{t-1}\bigg]\bigg]\\
    &\leq2\bb E_{\ezt}\bigg[\sum_{t=1}^T\bb E_{\ezt,\bfa\sigma}\bigg[\sup_{h_t\in\ca H_t}\sum_{j\in\ca V_t}\frac{1}{TN_t}\sigma_{t,j}L(h(X_{t,j}),Y_{t,j})\bigg|\pa Z_1^{t-1}\bigg]\bigg]\\
    &=2\bb E_{\ezt}\Vert\crad{L\circ\pa H}\Vert_1\\
    &\leq\bb E_{\ezt}\Vert\crad{\pa H}\Vert_1
\end{align*}
where the first inequality and the third inequality from the bottom comes from Jensen's inequality.  The second inequality from bottom comes from the symmetrization. The last inequality comes from the contraction lemma and the linearity of expectation.

\section{Proof of Theorem \ref{th:gen}}
By Chernoff bound and Azuma-Hoeffding's Inequality,
\begin{align*}
    &\bfa P\bigg(\lambda\sup_{\bfa h\in\pa H}\big(R_{\ezt}(\bfa h)-\widehat R_{Z_{\ca T}}(\bfa h)\big)-\lambda\bb E_{\pa Z_1^T}\big[ \sup_{\bfa h\in\pa H}\big(R_{\ca Z_{\ca T}}(\bfa h)-\widehat R_{Z_{\ca T}}(\bfa h)\big)\big]\leq \epsilon\bigg)\\
    &\leq\frac{\bb E_{\pa Z_1^T}\bigg[\exp\lambda\bigg(\sup_{\bfa h\in\pa H}\big(R_{\ezt}(\bfa h)-\widehat R_{Z_{\ca T}}(\bfa h)\big)-\bb E_{\pa Z_1^T}[\sup_{\bfa h\in\pa H}\big(R_{\ezt}(\bfa h)-\widehat R_{Z_{\ca T}}(\bfa h)\big)]\bigg)\bigg ]}{\exp(\lambda\epsilon)}\\
    &\leq\exp(-\lambda\epsilon)\bb E_{\pa Z_2^T}\bb E_{\pa Z_1}\bigg[\exp\lambda\bigg(\sup_{\bfa h\in\pa H}\big(R_{\ezt}(\bfa h)-\widehat R_{Z_{\ca T}}(\bfa h)\big)-\bb E[\sup_{\bfa h\in\pa H}\big(R_{\ezt}(\bfa h)-\widehat R_{Z_{\ca T}}(\bfa h)\big)]\bigg ]\\
    &\leq\exp(-\lambda\epsilon)\exp\bigg(\frac{\lambda^2}{2T^2|V_1|}\bigg)\bb E_{\pa Z_2^T}\bigg[\exp\lambda\bigg(\sup_{\bfa h\in\pa H}\big(R_{\ezt}(\bfa h)-\widehat R_{Z_{\ca T}}(\bfa h)\big)-\bb E_{\pa Z_1^T}[\sup_{\bfa h\in\pa H}\big(R_{\ezt}(\bfa h)-\widehat R_{Z_{\ca T}}(\bfa h)\big)]\bigg)\bigg|\pa Z_1\bigg ]\\
    &\leq \exp\bigg(\frac{\lambda^2}{2T^2}\big(\sum_{t=1}^T\frac{1}{N_t}\big)-\lambda\epsilon\bigg)
\end{align*}
Optimizing over $\lambda$, we know that the L.H.S is upperbounded by:
\begin{equation}
\label{la}
\exp\bigg(\frac{-2T^2\epsilon^2}{\sum_{t=1}^T\frac{1}{N_t}}\bigg)
\end{equation}
Using symmetrization and we introduce the Rademacher random field $\bfa\sigma=\{-1,+1\}^{\ca T}$ to get
\begin{align*}
\bb E_{Z_{\ca T}}\big[ \sup_{\bfa h\in\pa H}\big(R_{\ezt}(\bfa h)-\widehat R_{Z_{\ca T}}(\bfa h)\big)\big]&=\bb E_{Z_{\ca T}}[\sup_{\bfa h\in\pa H}\big(\bb E_{ Z_{\ca T}^\prime}\big(\widehat R_{Z_{\ca T}^\prime}(\bfa h)-\widehat R_{Z_{\ca T}}(\bfa h)\big)\big]\\
&\leq\bb E_{Z_{\ca T}, Z_{\ca T}^\prime}\big[\sup_{\bfa h\in\pa H}\big(\widehat R_{Z_{\ca T}^\prime}(\bfa  h)-\widehat R_{Z_{\ca T}}(\bfa h)\big)\big]\\
&=2\bb E_{\ezt}\bb E_{\bfa\sigma}[\sup_{h\in\pa H}\sum_{t=1}^T\frac{1}{TN_t}\sum_{j\in\ca V_t}L(h_t(X_{t,j}),Y_{t,j})]\\
&=2\bb E_{\ezt}\Vert\crad{L\circ\pa H}\Vert_1
\end{align*}
Since
\begin{equation*}
    L(h(x),y)=\frac{1-h(x)y}{2}
\end{equation*}
is $\frac{1}{2}$-Lipchitz w.r.t its first argument. 
We also have $\bb E_{\ezt}\Vert L\circ\crad{\pa H}\Vert_1 \leq\frac{1}{2}\bb E_{\ezt}\Vert\crad{\pa H}\Vert_1$ by contraction lemma and we complete the proof of the first inequality.

For the second inequality, it follows directly from the connection between empirical and Rademacher complexities that:
\begin{equation*}
    \Vert\ecrad{L\circ\pa H}\Vert_1\leq\Vert\crad{L\circ\pa H}\Vert_1+\frac{1}{T}\sqrt{\sum_{t=1}^T\frac{1}{2N_t}\log\frac{1}{\delta}}
\end{equation*}
And by union bound we obtain the second inequality.
\section{Proof of Theorem \ref{thm:ERMn}}
 Using the notation
 \begin{align*}
     \bfa h^*=\argmin_{\bfa h\in\pa H} R_\ezt(\bfa h)
 \end{align*}
By the definition of ERM, we immediately have the following:
 \begin{align*}
     R_{\ezt}(\bfa h_\text{ERM})-R_\ezt(\bfa h^*)&=R_\ezt (\bfa h_\text{ERM})-\wh R_\zt(\bfa h_\text{ERM})+\wh R_\zt(\bfa h_\text{ERM})-\wh R_\zt(\bfa h^*)\\
     &+\wh R_\zt(\bfa h^*)-R_\ezt(\bfa h^*)\\
     &\leq R_\ezt(\bfa h_\text{ERM})-\wh R_\zt(\bfa h_\erm)+\wh R_\zt(\bfa h^*)- R_\ezt(\bfa h^*)\\
     &\leq2\sup_{\bfa h\in\pa H}(R_\zt(\bfa h)-\wh R_\zt(\bfa h))
 \end{align*}
 By the uniform learn-ability, we have a rough estimate that
 \begin{align*}
     R_{\ezt}(\bfa h_\text{ERM})-R_\ezt(\bfa h^*)\leq2\Vert\ecrad{\pa H}\Vert_1+\frac{6}{T}\sqrt{\sum_{t=1}^T\frac{1}{2N_t}\log\frac{2}{\delta}}
 \end{align*}
 with probability at least $1-\delta$.
\section{Proof of Theorem \ref{ergodic}}
First, we divide index space $\ca T$ by its timestamps as $\{\ca V_1,...,\ca V_T\}$. Without loss of generality, we define $\ca V_0=\emptyset$.
Moreover, we let $\pa Z^{\setminus j}_t=(Z_{t,1},...,Z_{t,j-1},Z_{t,j+1},...,Z_{t,N_t})$ be the leave-one-out sample set at timestamp $t$.
For $j\in[N_t]$, we define 
\begin{equation*}
    V_{t,j}= \bb E [f(\pa Z_1^T)|\pa Z_1^{t-1}, Z_{t,j}]-\bb E [f(\pa Z_1^T)|\pa Z_1^{t-1}]
\end{equation*} with 
\begin{align*}
    f(\pa Z_1^T)&=[\sup_{\bfa h\in\pa H}[\wh R_{\zt}(\bfa h)-R_{\ezt}(\bfa h)]]-\bb E_{\ezt}[\sup_{\bfa h\in\pa H}[\wh R_{\zt}(\bfa h)-R_{\ezt}(\bfa h)]]
\end{align*}
being the empirical process on the uniform error deviation of $\pa H$.

We further show that
\begin{align*}
    V_{t,j} &=\bb E( f(\pa Z_1^T)|\pa Z_1^{t-1},Z_{t,j})-\bb E( f(\pa Z_1^T)|\pa Z_1^{t-1})\\
    &=\int_{z_{t}^{\setminus j},z_{t+1},...,z_T}\bfa P(dz_t^{\setminus j},dz_{t+1},..., dz_T|\pa Z_1^{t-1},Z_{t,j})\cdot f(\pa Z_1^{t-1},Z_{t,j},\pa Z_t^{\setminus j}=z_t^{\setminus j},\pa Z_{t+1}^T=z_{t+1}^T)\\
    &-\int_{z_{t},...,z_T}\bfa P(dz_{t},...,  dz_T|\pa Z_1^{t-1})\cdot f(\pa Z_1^{t-1},\pa Z_t^T=z_{t}^T)\\
    &\leq \sup_{z_{t,j}\in \bb Z}\int_{z_t^{\setminus j},...,z_T}\bfa P(dz_t^{\setminus j},dz_{t+1},..., dz_T|\pa Z_1^{t-1},Z_{t,j})\cdot f(\pa Z_1^{t-1},Z_{t,j}=z_{t,j},\pa Z_t^{\setminus j}=z_t^{\setminus j},Z_{t,j})\\
    &-\inf_{z_{t,j}\in \bb Z}\int_{z_t^{\setminus j},...,z_T}\bfa P(dz_t^{\setminus j},dz_{t+1},..., dz_T|\pa Z_1^{t-1},Z_{t,j})\cdot f(\pa Z_1^{t-1},Z_{t,j}=z_{t,j},\pa Z_t^{\setminus j}=z_t^{\setminus j},Z_{t,j})\\
    &:=M_{t,j}-m_{t,j}
\end{align*}
Moreover, using the assumption that the hypothesis class is $1$-Lipschitz, we see that
\begin{equation*}
   | f(\pa Z_1^T)-f(\pa Z_1^{\prime,T})| \leq\sum_{t=1}^T\sum_{j\in\ca V_t}\Vert Z_{t,j}-Z^\prime_{t,j}\Vert_1\cdot c_t 
\end{equation*}
with
\begin{equation*}
    c_t=\frac{1}{N_tT}
\end{equation*}
    
Then we define the Marton coupling with respect to the pair $(t,j)$ \begin{align*}
    (\pa Z_k,\pa Z_k^\prime) = (\pa Z_k^{(z_1,..., z_{t-1}, z_{t,j}, z_{t,j}^\prime)},\pa Z_k^{\prime (z_1,..., z_{t-1}, z_{t,j}, z_{t,j}^\prime)})
\end{align*}for all $j\in[t+1,T]$. We further let $z_t=(z_{t,1},...,z_{t,N_t})$ and $z_t^\prime=(z_{t,1},...,z_{t,j-1},z^\prime_{t,j},z_{t,j+1},...,z_{t,N_t})$ which differ from each other only at $j$. And we can see that
\begin{align*}
    M_{t,j}-m_{t,j}&=\sup_{z_1^{t-1},z_{t,j},z^\prime_{t,j}}\bigg[\bb E[f(\pa Z_1^T)|\pa Z_1^{t-1}=z_1^{t-1},Z_{t,j}=z_{t,j}]\\
    &-\bb E[f(\pa Z_1^T)|\pa Z_1^{t-1}=z_1^{t-1},Z^\prime_{t,j}=z^\prime_{t,j}]\bigg]\\
    &\leq\sup_{z_1^{t-1},z_{t,j},z^\prime_{t,j}}\bb E\bigg[\sum_{m=1}^T\sum_{k\in[N_t]}\Vert Z_{m,k}-Z^\prime_{m,k}\Vert_1c_t\bigg |\pa Z_1^{t-1}=\pa Z_1^{\prime t-1}=z_1^{t-1},Z_{t,j}=z_{t,j},Z_{t,j}^\prime=z^\prime_{t,j}\bigg]\\
    &\leq\sup_{z_1^{t-1},z_{t,j},z^\prime_{t,j}}\sum_{m=1}^T\sum_{k\in[N_t]}2TV\big (\bfa P(Z_{m,k}\big |\pa Z_1^{t-1}=\pa Z_1^{\prime t-1}=z_1^{t-1},Z_{t,j}=z_{t,j},Z_{t,j}^\prime=z^\prime_{t,j})\big )\\
    &\bfa P(Z^\prime_{m,k}|\pa Z_1^{t-1}=\pa Z_1^{\prime t-1}=z_1^{t-1},Z_{t,j}=z_{t,j},Z_{t,j}^\prime=z^\prime_{t,j}))c_t\\
    &=\sum_{k=t}^T 2\Gamma_{t,k}c_k
    \tag{*}\label{*}
\end{align*}
where the second inequality comes from the triangle inequality. Recall that
\begin{align*}
    \Gamma_{t,k}=\begin{cases}
     \sum_{j\in\ca V_t}\sup_{z_1^t,z_{t,j},z_{t,j}^\prime}TV(\bfa P(Z_{k,j}|\pa Z_1^t=z_1^t,Z_{t,j}=z_{t,j}),\bfa P(Z_{k,j}|\pa Z_1^t=z_1^t,Z_{t,j}=z_{t,j}^\prime)) & \text{if}\ k> t\\
      1,& \text{if}\ k= t\\
      0, & \text{otherwise}\\
    \end{cases}
\end{align*}
and for the mixing matrix we have

{\begin{align*}
     &\Gamma_{t,k}=\begin{cases}
    k_0\rho^{k-t} & \text{if}\ k\geq t\\
     1,& \text{if}\ k= t\\
      0, & \text{otherwise}\end{cases}
      \end{align*}}

Then we concluded by Azuma Hoeffding's inequality, observing that $\bb E[M_{t,j}-m_{t,j}|\pa Z_1^{t-1}]=0$ and note that $M_{t,j}-m_{t,j}\geq 0$, we have
\begin{align*}
    \bb E[\exp(\lambda V_{t,j})]\leq\bb E[\exp(\lambda(M_{t,j}-m_{t,j})]\leq\exp(\frac{\lambda^2}{2}(\sum_{k=t}^T\Gamma_{t,k}c_k)^2)
\end{align*}
which immediately leads to
\begin{align*}
    \bb E\bigg[\exp(\lambda f(\pa Z_1^T))\bigg]&=\bb E\bigg [\exp\bigg(\sum_{t=1}^T\sum_{j\in[N_t]}\lambda V_{t,j}\bigg)\bigg]\\
    &\leq\exp\bigg(\frac{\lambda^2}{2}\sum_{t=1}^T\sum_{j\in\ca V_t}\bigg( \sum_{k=t}^T\Gamma_{t,k}c_k\bigg)^2\bigg)\\
    &=\exp\bigg(\frac{\lambda^2}{2}\sum_{t=1}^TN_t\bigg( \sum_{k=t}^T\Gamma_{t,k}c_k\bigg)^2\bigg)\\
    &=\exp\bigg(\frac{\lambda^2}{2}\Vert\bfa N^{\frac{1}{2}}\bfa \Gamma\bfa c\Vert_2^2\bigg)
\end{align*}
with $\bfa c=(\frac{1}{N_1T},...,\frac{1}{N_TT})^\top$ and $\bfa N=diag\big(N_1,\ldots,N_T\big)$.
Hence, we conclude that, by the tail bound of sub-gaussian variables:
\begin{equation*}
    \bfa P\bigg([\sup_{\bfa h\in\pa H}[\wh R_{\zt}^g(\bfa h)-R_{\zt}^g(\bfa h)]]-\bb E_{\ezt}[\sup_{\bfa h\in\pa H}[\wh R_{\zt}^g(\bfa h)-R_{\zt}^g(\bfa h)]]\geq\epsilon\bigg)\leq\exp \big(-\frac{T^2\epsilon^2}{2\Vert\bfa N^{\frac{1}{2}}\bfa\Gamma\bfa c\Vert_2^2}\big)
\end{equation*}

The next step is to upperbound $\bb E_{\ezt}[\sup_{\bfa h\in\pa H}[\wh R_{\zt}^g(\bfa h)-R_{\zt}^g(\bfa h)]]$ by Rademacher complexity. Using the symmetrization, we have
\begin{align*}
    &\bb E_{\ezt}[\sup_{\bfa h\in\pa H}[\wh R_{\zt}^g(\bfa h)-R_{\zt}^g(\bfa h)]]\\
    &=\bb E_{\pa Z_1^T}\sum_{t=1}^T\bb E_{\pa Z_1^T}[\sup_{h\in\ca H_t}[\sum_{j\in\ca V_t}\frac{1}{TN_t}(L(h(X_{t,j}),Y_{t,j})-\bb E_{\pa Z_1^{t}}L(h(X_{t,j}),Y_{t,j}))]|\pa Z_1^{t-1}]\\
    &\leq2\bb E_{\pa Z_1^T}\Vert \crad{L\circ\pa H}\Vert_1
\end{align*}
We also note that the structural Rademacher complexity can also be bounded. First we note that
\begin{equation*}
    \Vert\ecrad{L\circ\pa H}|\pa Z^{\prime,T}_1\Vert_1-\Vert\ecrad{L\circ\pa H}|\pa Z_1^T\Vert_1\leq\sum_{t=1}^T\sum_{j\in\ca V_t}\Vert Z_{t,j}-Z^\prime_{t,j}\Vert_1\cdot \frac{c_t}{2}
\end{equation*}
Then we follow a similar argument as the concentration of measure and conclude that
\begin{equation*}
    \bfa P\bigg(\Vert\ecrad{L\circ\pa H}\Vert_1-\bb E_{\pa Z_1^T}[\Vert\crad{L\circ\pa H}\Vert_1]\geq\epsilon\bigg)\leq\exp\bigg(-\frac{\epsilon^2}{2\Vert\bfa N^{\frac{1}{2}}\bfa\Gamma\bfa c\Vert_2^2}\bigg)
\end{equation*}
Hence with probability at least $1-\delta$ the following holds:
\begin{equation*}
    \Vert\crad{L\circ\pa H}\Vert_1\leq\Vert\ecrad{L\circ\pa H}\Vert_1+\sqrt{2\log(\frac{1}{\delta})}\Vert\bfa N^{\frac{1}{2}}\bfa\Gamma\bfa c\Vert_2
\end{equation*}
Moreover, as the loss is $\frac{1}{2}$-Lipschitz with regard to the value of the function, by the contraction lemma ~\ref{contract}, we have:
\begin{equation*}
     \Vert\crad{L\circ\pa H}\Vert_1\leq\frac{1}{2} \Vert\crad{\pa H}\Vert_1\;\;\;\Vert\ecrad{L\circ\pa H}\Vert_1\leq\frac{1}{2} \Vert\ecrad{\pa H}\Vert_1
\end{equation*}
Substituting the structural Rademacher complexity with its empirical variant and we complete the proof.
\section{Proof of Theorem \ref{refine}}
The proof is built on a refinement over the mixing matrix. Recall in (\ref{*}) that:
\begin{align*}
    M_{t,j}-m_{t,j}
    &\leq \sum_{t=1}^T\sum_{j\in\ca V_t}2TV(\bfa P(Z_{t,j}|\pa Z_1^{t-1}=\pa Z_1^{\prime t-1}=z_1^{t-1},Z_{t,j}=z_{t,j},Z_{t,j}^\prime=z^\prime_{t,j}),\\
    &\bfa P(Z^\prime_{t,j}|\pa Z_1^{t-1}=\pa Z_1^{\prime t-1}=z_1^{t-1},Z_{t,j}=z_{t,j},Z_{t,j}^\prime=z^\prime_{t,j}))\\
    &=\sum_{k=t}^T 2N_t\Gamma_{t,k}c_k=4\sum_{k=t}^T\Gamma_{t,k}c
\end{align*}
In fact, this estimate is very rough as we have taken all $\ca V_k$ for $k\geq t$ into account over a single vertex $(t,j)\in\ca V_t$. In this refinement, we rewrite (\ref{*}) as:
\begin{align*}
    M_{t,j}-m_{t,j}&\leq\sum_{k=t}^T 2\ca N_{t,j}^k\frac{k_0}{N_k^2T}\rho^{k-t}
\end{align*}
This further suggests that
\begin{align*}
    \bb E\bigg(\exp(\lambda f(\pa Z_1^T))\bigg)
    &\leq\exp\bigg(\frac{\lambda^2}{2}\sum_{t=1}^T\sum_{j\in\ca V_t}\bigg( \sum_{k=t}^T \ca N_{t,j}^k\frac{k_0}{N_k^2T}\rho^{k-t}\bigg)^2\bigg)\\
\end{align*}
And using similar argument as the previous theorem, we have that:
\begin{align*}
    &\bfa P\bigg(\Vert\ecrad{L\circ\pa H}\Vert_1-\bb E_{\pa Z_1^T}[\Vert\crad{L\circ\pa H}\Vert_1]\geq\epsilon\bigg)\\
    &\leq\exp\bigg(-\frac{\epsilon^2}{2\sum_{t=1}^T\sum_{j\in\ca V_t}\big( \sum_{k=t}^T \ca N_{t,j}^k\frac{k_0}{N_k^2T}\rho^{k-t}\big)^2}\bigg)
\end{align*}
and
\begin{align*}
     &\bfa P\bigg([\sup_{\bfa h\in\pa H}[\wh R_{\zt}(\bfa h)-R_{\ezt}(\bfa h)]]-\bb E_{\ezt}[\sup_{\bfa h\in\pa H}[\wh R_{\zt}(\bfa h)-R_{\ezt}(\bfa h)]]\geq\epsilon\bigg)\\
     &\leq\exp \bigg(-\frac{\epsilon^2}{2\sum_{t=1}^T\sum_{j\in\ca V_t}\big( \sum_{k=t}^T \ca N_{t,j}^k\frac{k_0}{N_k^2T}\rho^{k-t}\big)^2}\bigg)
\end{align*}
By substituting the r.h.s. of above two inequality with $\delta/2$ and use union bound to combine them together, we complete the proof.
\section{Proof of Theorem \ref{thm:34}}
The proof follows a similar discussion as the non-communicative case. Where the expected regret of ERM, non-asymptotically, can be upper bounded by two times the maximum deviation of empirical processes. However, we note that the structural Rademacher complexity became a process itself and the bound is taking its expectation.

\end{document}